\documentclass[lettersize,journal]{IEEEtran}

\ifCLASSINFOpdf
\else
   \usepackage[dvips]{graphicx}
\fi
\usepackage{caption}
\usepackage{tabularx}
\usepackage{multirow}
\usepackage{multicol}
\usepackage{makecell}
\usepackage{graphicx}
 \usepackage{hyperref}
\usepackage{amssymb}
\usepackage{booktabs}
\usepackage{amsmath}
\usepackage{algorithm}
\usepackage{algorithmic}
\usepackage{textcomp}
\usepackage{stfloats}
\usepackage{url}
\usepackage{xcolor}
\usepackage{verbatim}
\usepackage{cite}
\usepackage{subfigure}
\usepackage{listings}

\newcommand{\tableref}[1]{\hyperref[#1]{\textcolor{blue}{Table }\textup{(\ref{#1})}}}
\newcommand{\figref}[1]{\hyperref[#1]{\textcolor{blue}{Fig.}\textup{(\ref{#1})}}}
\newcommand{\equref}[1]{\hyperref[#1]{\textcolor{blue}{Eq.}\textup{\ref{#1}}}}
\begin{document}

\title{Precipitation Nowcasting Using Diffusion Transformer with Causal Attention}

\author{
ChaoRong Li*, 
\IEEEmembership{Member, IEEE}, XuDong Ling*, YiLan Xue, Wenjie Luo, LiHong Zhu, FengQing Qin,  Yaodong Zhou,Yuanyuan Huang
\thanks{ This work was supported in part by the Science and Technology Department of Sichuan Province under Grant 2023YFQ0011, in part by Yibin University under Grant 2023YY02.}
\thanks{C.R. Li is with the Faculty of Artificial Intelligence and Big Data, Yibin University, Yibin 644000,China and Key Laboratory of Urban-Rural Industrial Integration and Intelligent Decision-Making, Sichuan Province For Social Sciences (e-mail: lichaorong88@163.com).}
\thanks{X.D. Ling and W.J Luo are with the Faculty of Artificial Intelligence and Big Data, Yibin University, Yibin 644000,China
and Chongqing University of Technology, Chongqing 400054,China (e-mail: clearlyzero@stu.cqut.edu.cn, lwj018@stu.cqut.edu.cn).}

\thanks{L.H Zhu, F.Q. Qin, Y.L Xue and Y.D Zhou are with the Faculty of Artificial Intelligence and Big Data, Yibin University, Yibin 644000,China (e-mail:zhulihong2022@student.usm.my,qinfengqing16@163.com,
xuemomode@foxmail.com,
1170140118@qq.com).

(\textit{Corresponding author:C.R. Li, *indicates equal contribution}).}
\thanks{Yuanyuan Huang is with the Chengdu University of Information Technology,Chengdu 610225,China(e-mail: hy@cuit.edu.cn).}

}

\maketitle

\begin{abstract}
	Short-term precipitation forecasting remains challenging due to the difficulty in capturing long-term spatiotemporal dependencies. Current deep learning methods fall short in establishing effective dependencies between conditions and forecast results, while also lacking interpretability. To address this issue, we propose a Precipitation Nowcasting Using Diffusion Transformer with Causal Attention model.
Our model leverages Transformer and combines causal attention mechanisms to establish spatiotemporal queries between conditional information (causes) and forecast results (results). This design enables the model to effectively capture long-term dependencies, allowing forecast results to maintain strong causal relationships with input conditions over a wide range of time and space.
We explore four variants of spatiotemporal information interactions for DTCA, demonstrating that global spatiotemporal labeling interactions yield the best performance. In addition, we introduce a Channel-To-Batch shift operation to further enhance the model's ability to represent complex rainfall dynamics. We conducted experiments on two datasets. Compared to state-of-the-art U-Net-based methods, our approach improved the CSI (Critical Success Index) for predicting heavy precipitation by approximately 15\% and 8\% respectively, achieving state-of-the-art performance.

\end{abstract}


\IEEEpeerreviewmaketitle

\section{Introduction}
\IEEEPARstart{N}owcasting, the youngest yet most promising branch of weather forecasting, primarily focuses on rainfall predictions within the next 12 hours. It is particularly adept at forecasting small to medium-scale weather systems with short lifespans and sudden onset, such as severe convective storms. By providing forecasts 1-2 hours in advance, nowcasting can accurately capture the location of weather systems and predict rainfall areas, enabling timely communication of rainfall information and alerting the public to take necessary precautions \cite{ravuri2021skilful,sit2024efficientrainnet}.

Numerical Weather Prediction (NWP) has long been a powerful tool for generating short-term forecasts using atmospheric initial conditions \cite{chirigati2021accurate}. However, NWP systems face significant challenges in accurately describing the evolution of convective precipitation systems at short spatiotemporal scales due to insufficient initial conditions and computational resource limitations. This makes it difficult to precisely predict local weather phenomena.

Given these limitations of NWP, researchers have turned to alternative methods for short-term precipitation forecasts ranging from minutes to hours. These methods typically employ high-resolution real-time precipitation data observed by weather radar. Extrapolation techniques, such as optical flow advection schemes \cite{bowler2006steps,pulkkinen2019pysteps}, are used to predict the future movement direction and velocity of rain bands, aiming to achieve accurate estimates of precipitation intensity, affected areas, and duration.
Nevertheless, extrapolation methods have their own limitations. The underlying stationarity assumption, which presumes that the current state of the atmosphere remains unchanged, and the simplified treatment of the flow field restrict their performance in short-term forecasting. These limitations become increasingly prominent when facing the complex nonlinear characteristics of precipitation evolution as the forecast lead time increases \cite{skillful}.
  
In recent years, deep learning technology has been leading technological innovation and has shown outstanding capabilities in simulating complex Earth systems and addressing Earth science challenges, such as short-term precipitation forecasting \cite{narkhede2022review,zhang2023skilful,skillful}. In precipitation forecasting tasks, the methods used can be divided into non-generative and generative prediction methods. Non-generative prediction treats precipitation forecasting as a deterministic spatiotemporal prediction task. These methods train networks to predict future precipitation by performing supervised learning on historical and future weather radar image sequences, with the objective function being a distance function or a classification objective function. In recent years, non-generative precipitation forecasting has made significant progress in network architecture. An increasing number of studies have introduced advanced model architectures into precipitation forecasting, such as Convolutional Neural Networks (CNNs), Recurrent Neural Networks (RNNs), and Transformers \cite{vaswani2017attention}, to build models with better long-term dependency and global perception capabilities. For example, Shi et al. combined convolutional networks and RNNs and proposed Conv-LSTM \cite{shi2015convolutional} and Conv-GRU \cite{shi2017deep} models, aiming to effectively extract and utilize the spatiotemporal correlations in sequential radar data to improve the accuracy of precipitation forecasting. Ning et al. \cite{ning2023mimo} proposed MIMO-VP, a multi-input multi-output Transformer-based video prediction architecture that has achieved good results in long-term prediction. However, models trained based on these simple loss functions can only reflect the average difference between the predicted results and the true values, resulting in overly smooth and blurry predictions \cite{trebing2021smaat} that do not align with human perception. Moreover, they fail to capture the potential diversity and uncertainty in the predicted results \cite{skillful}, which greatly limits the application of deep learning in hydrological analysis. 

To address these limitations, particularly the issue of blurry predictions in nowcasting, generative methods have emerged as an effective solution. These methods train deep models using existing data to learn and understand data distributions, then perform conditional sampling within the learned distribution to generate accurate short-term nowcasts. This approach not only produces more realistic prediction results but also demonstrates better overall capabilities in handling extreme weather events.
Numerous studies have explored nowcasting precipitation using generative models, including Variational Autoencoders (VAEs), flow models (FLOW), Generative Adversarial Networks (GANs), and Diffusion Models (DMs). For instance, Ravuri et al. \cite{skillful} proposed a GAN with regularization terms, called DGMR. The rainfall results generated by this network are not only numerically precise but also highly similar to real rainfall fields in terms of shape and motion trajectory. Zhang et al. \cite{harris2022generative} introduced NowcastNet, a nonlinear probabilistic generative model for extreme precipitation nowcasting. They incorporated physics-compliant evolution networks into their model, addressing the issue of data-driven learning methods not adhering to physical laws. Ling et al. \cite{10480402} proposed TSDM (Two-stage Rainfall Forecasting Diffusion Model) based on diffusion models, which is a two-stage rainfall prediction method. This approach first conducts spatiotemporal prediction at low resolution and then uses a diffusion model for spatial super-resolution restoration, achieving good results in both spatial and temporal aspects.

The U-Net architecture, widely used in image tasks including weather forecasting, has limitations for complex spatiotemporal tasks like precipitation nowcasting. Its upsampling and downsampling operations lose crucial fine-grained details. U-Net's structure fails to fully capture long-range dependencies in large-scale weather patterns, and its fixed receptive field adapts poorly to varying scales of weather phenomena. The Transformer \cite{vaswani2017attention} architecture has become the preferred model for various tasks due to its outstanding performance. In the field of image generation, compared to the alternating upsampling and downsampling operations in the U-Net architecture, the end-to-end design of the Transformer can more effectively retain the detail features of the input. This advantage is attributed to the self-attention mechanism of the Transformer, which enables it to capture long-range dependencies in the input sequence and preserve the original information to the greatest extent. Based on this, Peebles et al. \cite{peebles2023scalable} proposed DIT, successfully replacing the U-Net architecture with the Transformer architecture. This architecture eliminates the upsampling and downsampling operations in U-Net and, through the self-attention mechanism, can better capture long-range dependencies and retain more detail information, thereby improving the quality of generated images. Furthermore, the Sora model \cite{videoworldsimulators2024} proposed by OpenAI utilizes a Transformer architecture with spatiotemporal patches, acting on the latent encoding of videos and images to capture correlations, and can generate long-duration, high-resolution videos based on text prompts while adhering to the physical laws of the real world.

Despite the advantages of Transformer architectures in capturing long-range dependencies and preserving detailed information, most models in the field of short-term precipitation forecasting still employ the U-Net architecture as the backbone network. Additionally, few studies have effectively incorporated conditional information, such as historical weather data or geographical features, which could significantly improve the accuracy of precipitation prediction. To address these challenges and leverage the strengths of Transformer architectures, we propose the Diffusion Transformer with Causal Attention (DTCA) method. This novel approach aims to significantly enhance the performance of precipitation forecasting by combining the advantages of Transformer architectures with effective integration of conditional information.

Our main contributions are as follows:
\begin{itemize}
\item  We introduce a novel  causal attention mechanism based on conditional precipitation distribution feature-related queries, which can effectively capture complex spatiotemporal precipitation patterns and significantly improve prediction performance.
\item We propose an innovative spatiotemporal Transformer diffusion model and systematically explore and compare various spatiotemporal feature capture structures. We find that the Full Joint Time-Space attention mechanism excels in capturing spatiotemporal dependencies.
\item We propose the Channel-To-Batch Shift operation, combined with the causal attention mechanism, to enhance the model's understanding and representation capabilities for complex precipitation patterns, effectively improving prediction accuracy. Moreover, we analyze the impact of different shift amounts on the accuracy of precipitation prediction.
\end{itemize}

\begin{figure*}[!t] 
	\centering	
	\includegraphics[width = \linewidth]{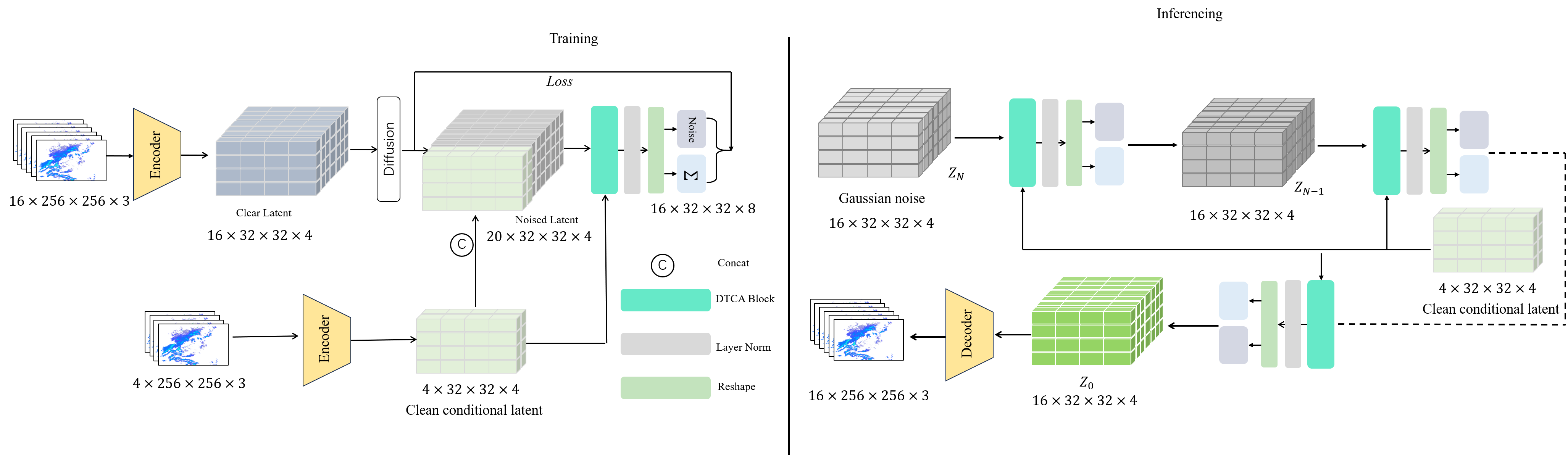}
	\caption{ Architecture diagram}
	\label{t_p}
  \end{figure*}

  \section{Related Works}
  \subsection{Diffusion Model}
  Diffusion models(DM) \cite{ho2020denoising}, as an emerging type of image generation model, have achieved significant success in recent years. Compared to traditional generative adversarial networks (GANs) \cite{goodfellow2014generative} and variational autoencoders (VAEs) \cite{kingma2013auto}, diffusion models have demonstrated superior performance in many vision-related tasks \cite{ho2022classifier,ho2022imagen,lu2022dpm}. Unlike GANs, the training process of diffusion models is more stable, without the issue of mode collapse, and the generated images have higher quality and no artifacts. Compared to VAEs, the images generated by diffusion models not only have higher quality but also exhibit greater diversity. These advantages have made diffusion models an important breakthrough in the field of image generation, opening up new possibilities for related applications.
Diffusion models consist of two Markov chain processes: noise addition and denoising. The forward process has a total of T=1000 steps, slowly adding noise to the real data $X_0$ and generating a series of noisy samples $X_1, X_2, ..., X_T$. Each step of the noise addition process can be represented as \equref{difffor}, during which the data samples gradually lose their original features until they completely transform into Gaussian noise.

\begin{align}
	\label{difffor}
  q(x_t|x_t-1) = N(x_t;\sqrt{\alpha}x_{t-1},(1-\alpha_t)I)
  \end{align}
By applying the reparameterization trick, we can sample $x_t$ as shown in \equref{difffor1}:
\begin{align}
	\label{difffor1}
	x_t= \sqrt{\bar{\alpha}_t}  x_0 + \sqrt{1 - \bar{\alpha}_t} \epsilon_t 
  \end{align}

  where $\bar{\alpha}_t = \prod ^t _{i=1}\alpha_i $,$\alpha_t = 1-\beta _t$,$\beta _t \in (0,1)^T _t=1$

The reverse process of diffusion models aims to reverse the noise introduced in the forward process to recover the original data. In diffusion models, the goal of the reverse process is to recover the original data $x_{t-1}$ from the noise-corrupted data $x_t$. Typically, the reverse process is achieved by training a reverse model that takes the noise-corrupted data $x_t$ as input and outputs the distribution of the recovered original data $x_{t-1}$ (as shown in \equref{backword}). The training objective of the reverse model is to minimize the difference between the predicted noise $\epsilon_\theta(x_t)$ and the real noise $\epsilon$, i.e., $\bigtriangledown_{\theta}|| \epsilon -\epsilon_\theta(x_t)||^2$, where $\epsilon_\theta$ represents the network structure of the reverse model. By optimizing this objective function, the reverse model can learn how to effectively recover the original information from the noise-corrupted data, thereby achieving high-quality data generation.

\begin{align}
	\label{backword}
  X_{t-1} = \frac{1}{\sqrt{\alpha_t}}(X_t -1\frac{1-\alpha_t
  }{\sqrt{1-\bar{\alpha}_t}}) 
  \end{align}
  
  \subsubsection{Latent Diffusion Model}
  The training and inference computational costs of diffusion models in pixel space are extremely high. To address this issue, Rombach et al. proposed the Latent Diffusion Model (LDM) \cite{rombach2022high}, which employs a two-stage approach. In the first stage, an autoencoder is trained to learn how to compress high-dimensional images into a low-dimensional latent space and use a decoder to restore the original images. In the second stage, the pre-trained Encoder (at this stage, the Encoder and Decoder parameters are frozen and no longer undergo gradient updates) is used to map the images to the latent space $Z = E(x)$, and then a diffusion model is applied in the latent space to learn the distribution of the latent variables. Finally, the latent variables are decoded into high-quality images through the decoder $X=D(z)$. This approach effectively reduces the computational complexity, enabling diffusion models to be applied to larger-scale image generation tasks.

  \subsection{Transformer for Diffusion Model}

  In the field of image generation, researchers have proposed various Transformer-based models and made significant progress. Bao et al. \cite{bao2023all} proposed U-ViT based on the Vision Transformer (ViT)\cite{dosovitskiy2020image} and successfully applied it to image generation tasks using diffusion models, achieving effects comparable to convolutional neural networks (CNNs) of the same scale. Peebles et al. \cite{peebles2023scalable} proposed a novel Transformer model called DIT that introduces an Adaptive Layer Normalization module as a regulatory mechanism, which has achieved excellent performance on the class-conditional ImageNet generation task. Furthermore, the PixArt model proposed by Chen et al. \cite{chen2023pixart} further improves computational efficiency by modifying the conditional mapping approach of the DIT Block. The design goal of PixArt is to significantly reduce training costs while maintaining generation quality comparable to the current state-of-the-art image generators.


\section{Method}
\subsection{Framework Overview}
Our proposed rainfall prediction framework, illustrated in \figref{t_p}, builds upon the Encoder-Decoder structure introduced in \cite{ling2024spacetime}. This framework maps the input from pixel space to latent space, employs a neural network to predict future precipitation scenarios, and then decodes the results back to pixel space. To accurately describe the input data's dimensional information, we define its shape as $Z \in \mathbb{R}^{B \times C \times F \times N}$, where $B$ represents the batch size, $C$ the number of tokens (channels), $F$ the number of time frames, and $N$ the length of a single token. In the diffusion model for predicting noise, we adopt a strategy of concatenating the condition tokens and noise tokens along the time dimension ($F$), forming input data with dimensions $Z \in \mathbb{R}^{B \times C \times (F_c+F_n) \times N}$. Here, $F_c=4$ represents the number of condition frames and $F_n=16$ the number of noise frames to be predicted. After forward propagation through the model, we extract the last $F_n=16$ frames from the output as the prediction result, obtaining $Z_{out} \in \mathbb{R}^{B \times C \times F_n \times N}$. This processing approach allows for full utilization of both condition and noise information while ensuring that the model output has the same dimensions as the target true rainfall sequence. Once the denoising process is complete, the decoder maps the predicted latent space back to the pixel space, concluding the entire prediction process. By structuring the framework in this way, we effectively leverage historical data (condition frames) to predict future rainfall scenarios while maintaining dimensional consistency throughout the process.
\begin{figure*}[!t] 
	\centering	
	\includegraphics[width = \linewidth]{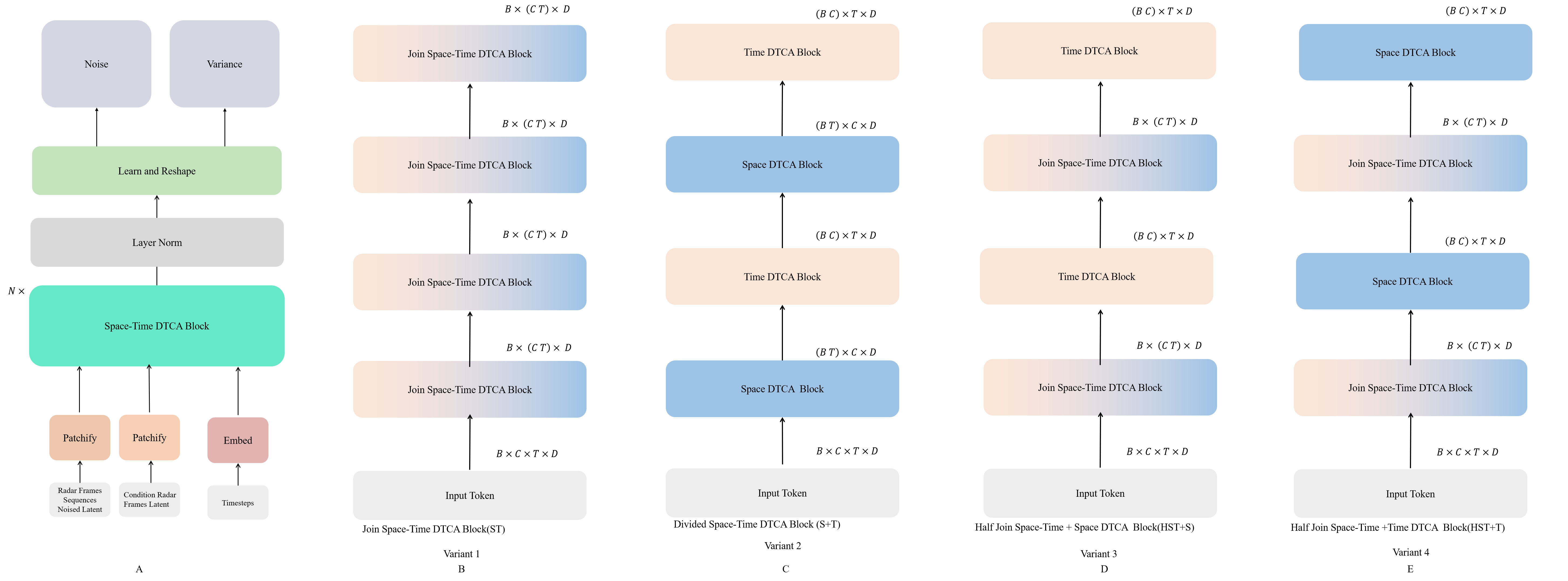}
	\caption{DTCA architecture diagram}
	\label{fff}
  \end{figure*}
\begin{figure}[!t] 
  \centering	
  \includegraphics[width = 0.6\linewidth]{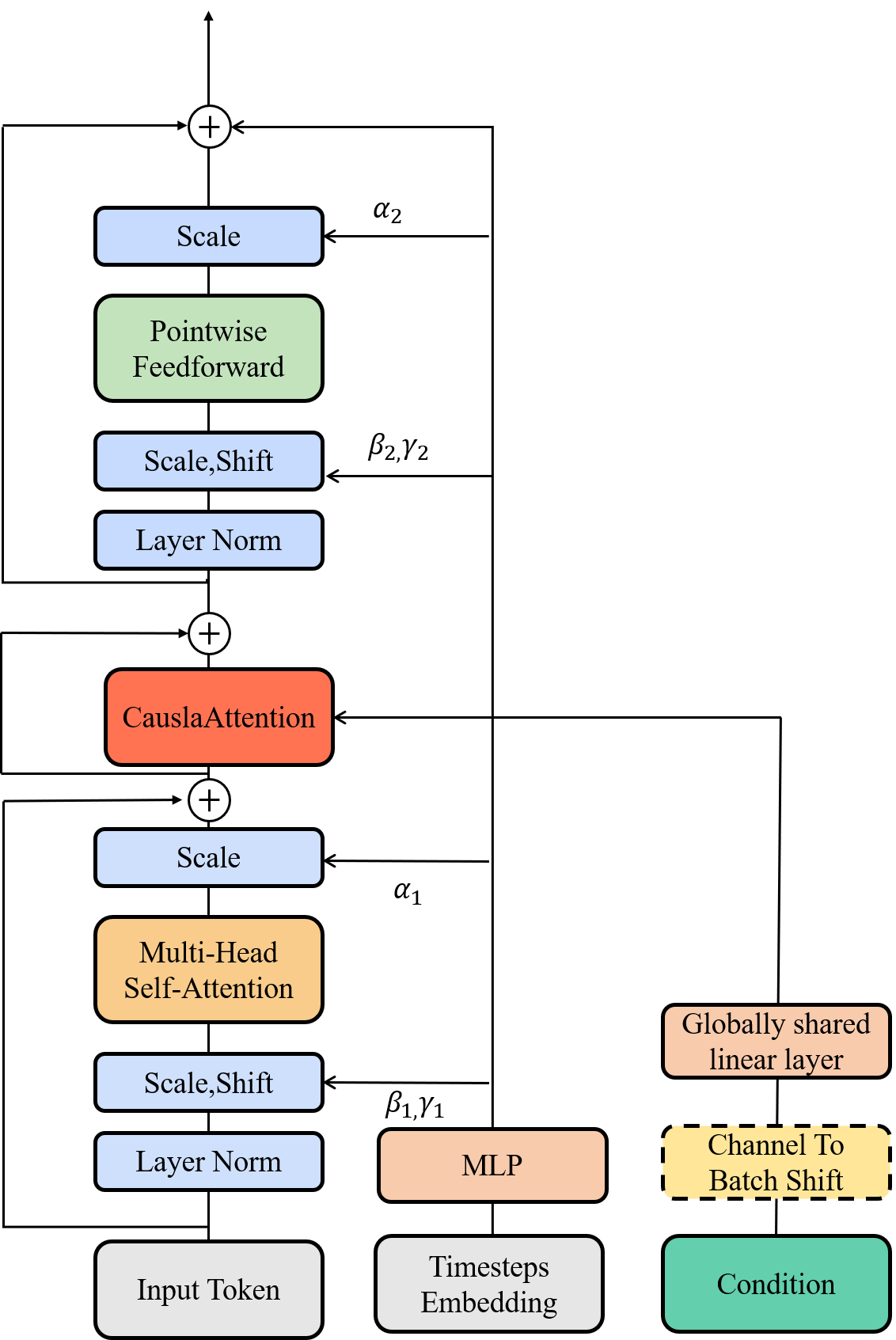}
  \caption{DTCA Block architecture diagram}
  \label{vlock}
  \end{figure}
\begin{figure*}[!t] 
  \centering	
  \includegraphics[width = \linewidth]{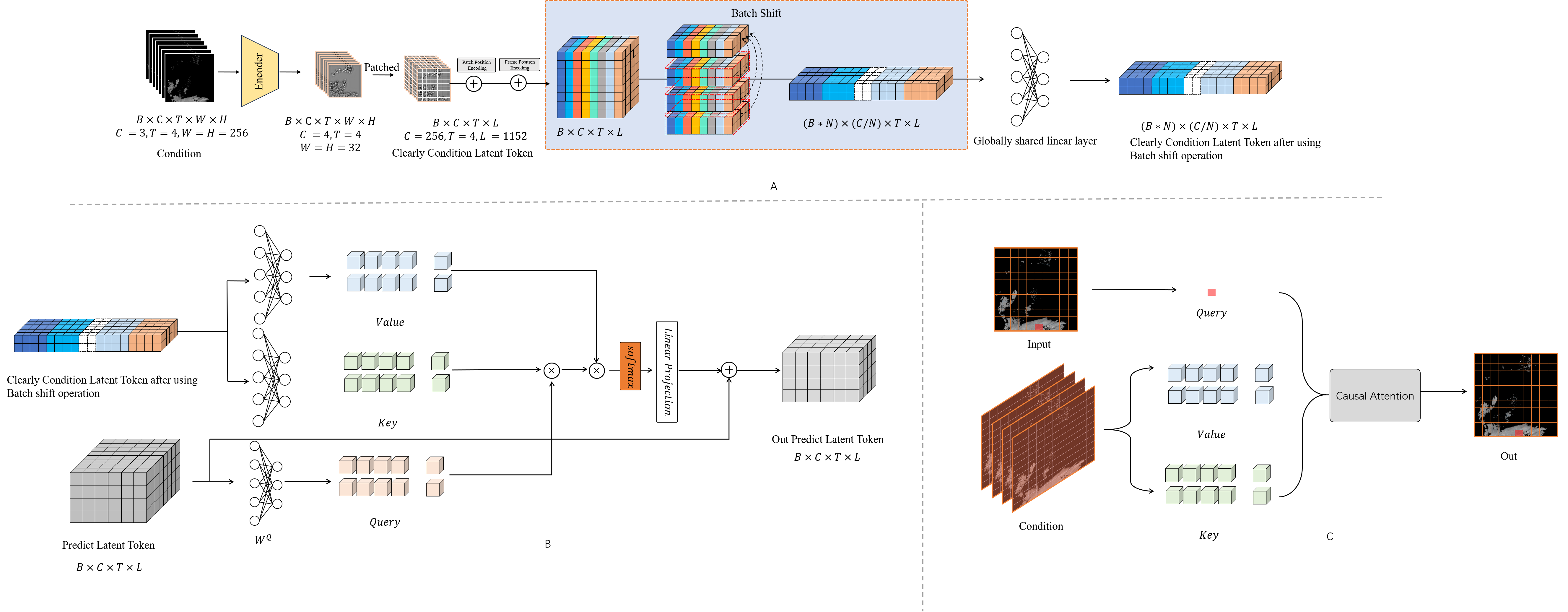}
  \caption{Channel-To-Batch Shift Causal Attention operation diagram}
  \label{f4}
\end{figure*}

\subsection{Spatio-Temporal Patch Extraction}
\label{patch}
The spatial resolution of our rainfall image sequence is $256 \times 256$, with a time span of 20 frames (100 minutes). First, we use AutoEncoder-KL\cite{ling2024spacetime} to compress the rainfall sequence images from the original size of $F=20 \times W=256 \times H=256 \times C=1$ to a latent space size of $F=20\times W=32 \times H=32 \times C=4$. Then, we divide the latent space into $2 \times 2$ sized patches \cite{dosovitskiy2020image}, mapping each patch to a vector of length 1152, ultimately obtaining spatiotemporal tokens with a shape of $F=20 \times C=256 \times N=1152$, where the first 4 frames serve as conditions and the latter 16 frames are the sequence to be predicted.

\subsection{The Model variants of DTCA}
Our proposed DTCA spatiotemporal denoising model is shown in \figref{fff}. The model's input includes noise, the first 4 frames of observed radar image sequences, and the diffusion time T. The core architecture of the model consists of modules that process spatiotemporal data of different dimensions (as shown in \figref{vlock}). The DTCA module effectively captures the complex spatiotemporal dependencies between input tokens through a multi-head self-attention mechanism, and uses modulation technology to dynamically introduce diffusion moment information into the model. DTCA also combines \textbf{Causal Attention} and \textbf{Channel-To-Batch Shift (CTBS)} to introduce conditional information to ensure the causal consistency of the generation process. To thoroughly investigate how to effectively capture and process spatiotemporal data in rainfall prediction, we designed four variant methods for processing spatiotemporal data.

\textit{\textbf{Variant 1}}: Full Join Space-Time DTCA Block,This variant focuses on jointly processing the space-time dimension to fully perceive the changes in the space-time dimension. In this design, $Z$ is reshaped into $Z \in \mathbb{R}^{B \times (C F) \times N}$, enabling the model to capture the correlation features in space and time at the same time.

\textit{\textbf{Variant 2}}:Divided Space-Time DTCA Block, in the Divided Space-Time DTCA Block, we capture spatiotemporal information by processing spatial and temporal dimensions separately. Specifically, the initial input token shape is $Z \in \mathbb{R}^{B \times C \times F \times N}$. First, we reshape $Z$ into $Z \in \mathbb{R}^{(B F) \times C \times N}$ and input it into the spatial DTCA Block. This step allows the module to accurately compute spatial correlations, significantly enhancing its ability to process spatial features. Then, we further reshape $Z$ into $Z \in \mathbb{R}^{(B C) \times F \times N}$ to capture the dynamic changes in the time series.

\textit{\textbf{Variant 3}}: Half Join Space-Time \& Space DTCA Block. This is a variant that combines the Full Join Space-Time and Divided Space-Time methods. First, the model reshapes the input $Z$ into $Z \in \mathbb{R}^{B \times (C F) \times N}$ and inputs it into the Full Join Space-Time DTCA module to jointly process the space-time dimensions. Subsequently, to further refine spatial information, we reshape $Z$ into $Z \in \mathbb{R}^{(B F) \times C \times N}$ and input it into the spatial DTCA module for spatial modeling.

\textit{\textbf{Variant 4}}: Half Join Space-Time \& Time DTCA Block
This variant adopts a design similar to the Half Join Space-Time \& Space DTCA Block, but focuses on modeling the temporal dimension in the second stage. Specifically, the model first reshapes the input $Z$ into $Z \in \mathbb{R}^{B \times (C F) \times N}$ and inputs it into the Joint Space-Time DTCA module to learn combined spatiotemporal features. Then, to capture the dynamic evolution of the time series, we further reshape $Z$ into $Z \in \mathbb{R}^{(B C) \times F \times N}$ and input it into the temporal DTCA module.

\begin{figure*}[!t]
  \centering
  \subfigure[Token Similarity between Samples along Spatial Dimension on SW Dataset]{
  \includegraphics[width=0.48\textwidth]{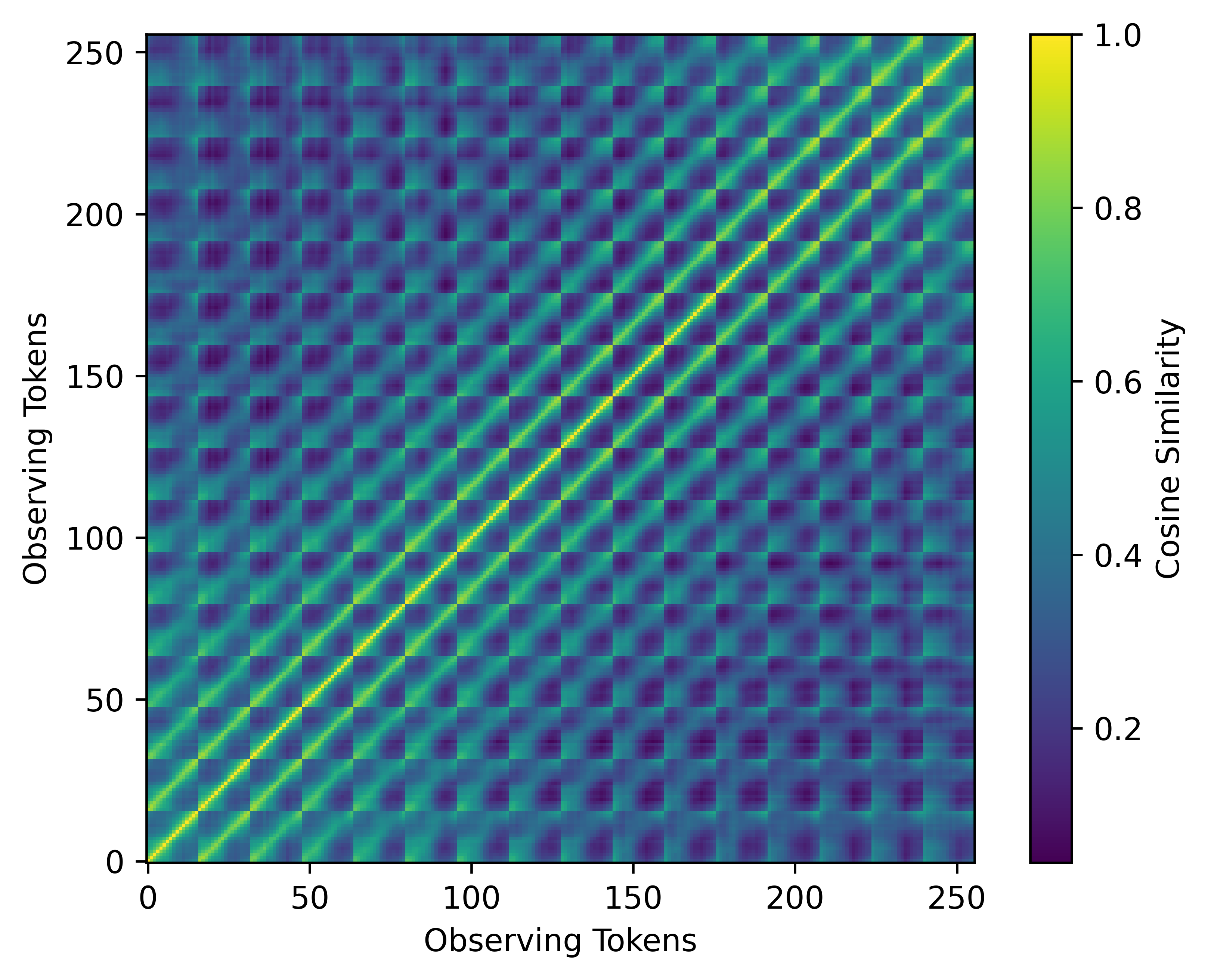}
  \label{token_1self}
  }
  \hfill
  \subfigure[Token Similarity between Samples along Spatial Dimension on MRMS Dataset]{
  \includegraphics[width=0.48\textwidth]{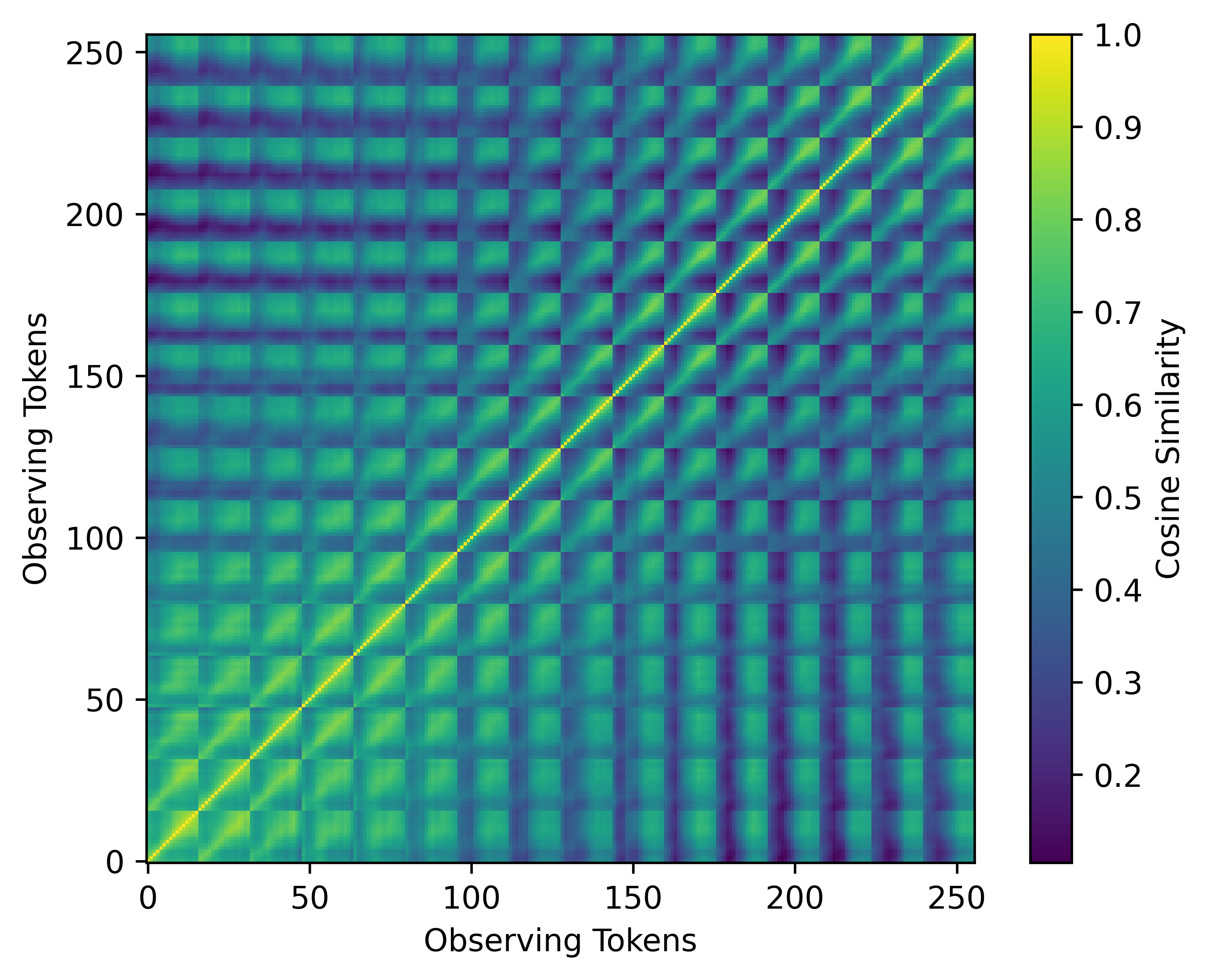}
  \label{token_2self}
  }

  \subfigure[Token Similarity between Samples along Temporal Dimension on SW Dataset]{
    \includegraphics[width=0.48\textwidth]{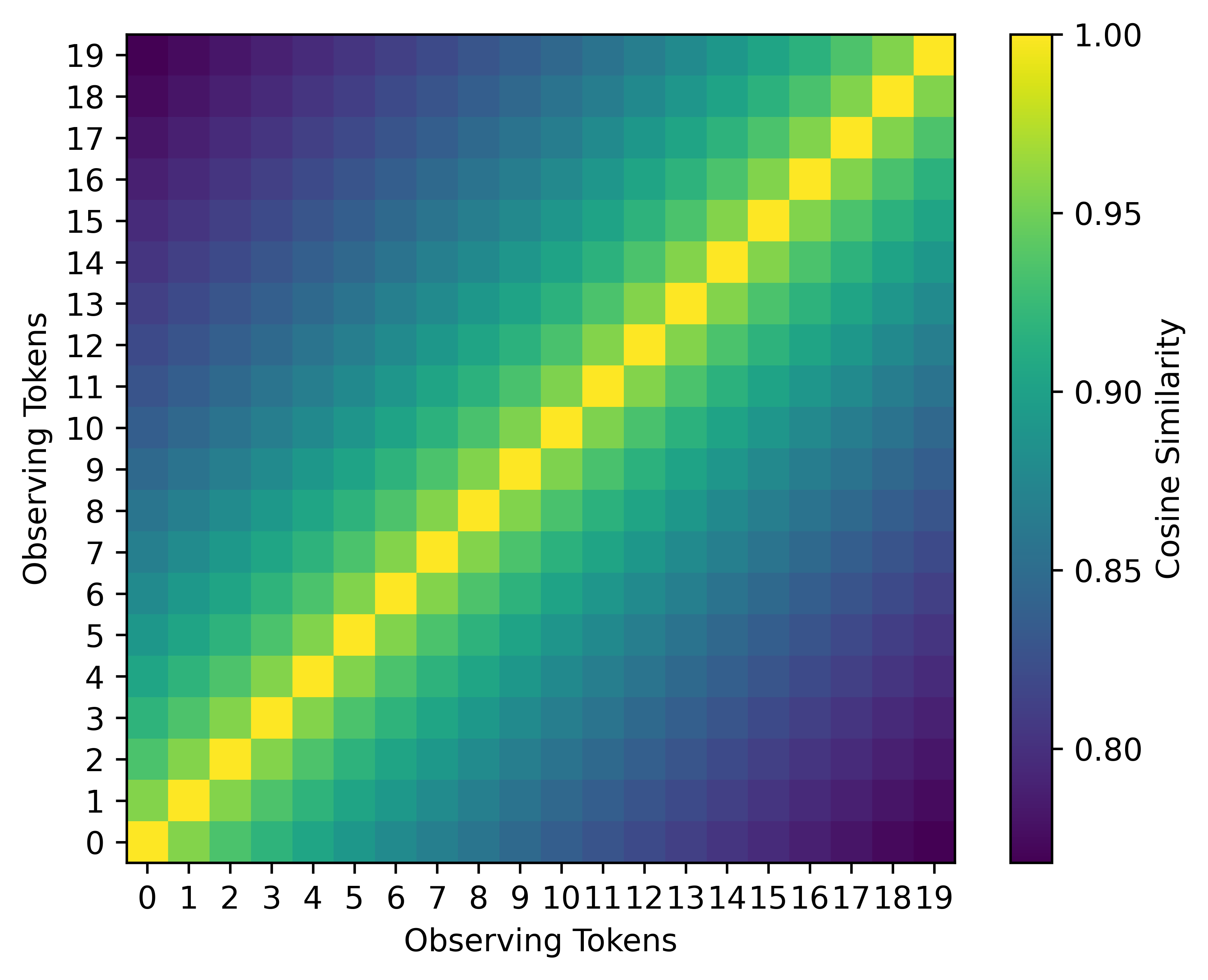}
    \label{token_1self1}
    }
    \hfill
    \subfigure[Token Similarity between Samples along Temporal Dimension on MRMS Dataset]{
    \includegraphics[width=0.48\textwidth]{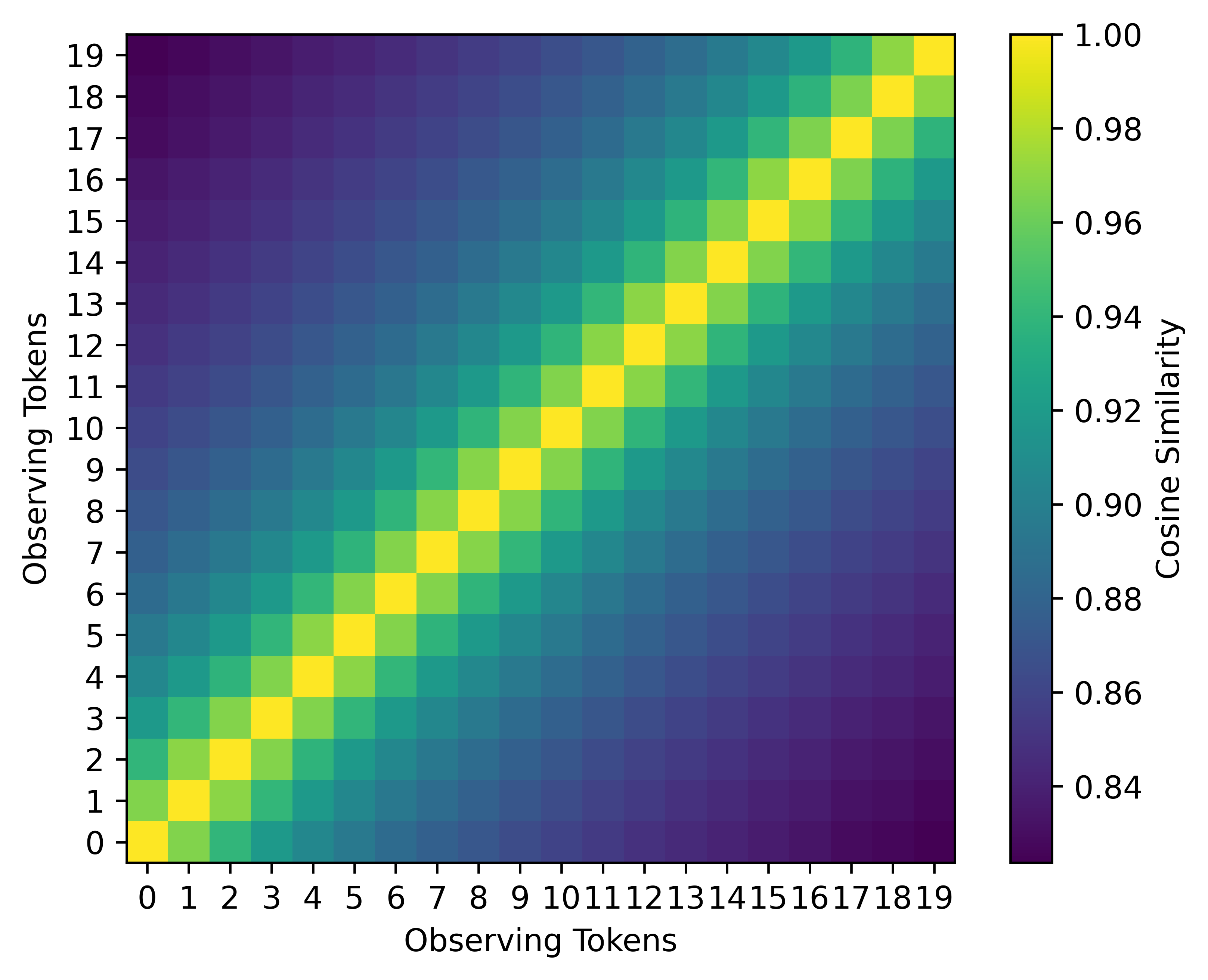}
    \label{token_2self1}
    }

  \caption{Token similarity visualization along different dimensions.}
  \label{token_self_sw}
  \end{figure*}

  \begin{algorithm}[t]
    \caption{Channel-to-Batch Shift Causal Attention}
    \label{alg:code}
    \definecolor{codeblue}{rgb}{0.25,0.5,0.5}
    \lstset{
      backgroundcolor=\color{white},
      basicstyle=\fontsize{7.2pt}{7.2pt}\ttfamily\selectfont,
      columns=fullflexible,
      breaklines=true,
      captionpos=b,
      commentstyle=\fontsize{7.2pt}{7.2pt}\color{codeblue},
      keywordstyle=\fontsize{7.2pt}{7.2pt},
      texcl=<true|false>
    }
    \begin{lstlisting}[language=python]
    # X: input of shape (B,(C T),N)
    # Condition: input of shape (B,C,T,N)
    # Frame_pos : Frame position information
    # Patch_pos : Ptach location information
  
    Condition = rearrange(Condition,"(B,C,T,N)->(B,T),C,N")+Patch_pos #Add patch location information
    Condition = rearrange(Condition,"(B,T),C,N->((B,C),T,N)") #Add frame position information
    Condition = rearrange(Condition,"(B,C),T,N->((B*N),(C/N),N"))#Channel-To-Batch Shift core operation, N represents the number of shifts 
    Condition = nn.Linear(Condition)# Use global conditional linear layers to prevent gradient vanishing
    # In one of the space-time blocks
        ...#other operations
        X = CausalAttention(X,Condition)
        ...#other operations
  
  
  
  
  
    \end{lstlisting}
    \end{algorithm}

\subsection{Causal Attention and Channel-To-Batch Shift}
\label{cacts32}
In this study, we observed 20 frames of rainfall fields from the entire dataset, including 4 frames of conditional data and 16 frames of data to be predicted. Based on the spatiotemporal token generation method described in Section \ref{patch}, we generated a total of $256 \times 20$ tokens in the spatiotemporal dimension. To conduct an in-depth analysis of the characteristics and correlations of the rainfall fields in the spatiotemporal dimension, we calculated the cosine similarity of these tokens in both spatial and temporal dimensions. The results are shown in \figref{token_self_sw}.

From a spatial perspective, the entire similarity matrix exhibits a distinct block structure. This phenomenon indicates that the spatial tokens can be divided into several highly correlated regions or sub-blocks. Tokens within each sub-block have high similarity to each other, while the similarity between different sub-blocks is relatively low. Notably, the size and shape of the sub-blocks are not entirely uniform, suggesting the spatial inhomogeneity of the rainfall field. Some larger sub-blocks indicate areas where rainfall distribution is relatively uniform, while smaller sub-blocks correspond to local rainfall centers or anomalous regions. Even in areas far from the diagonal, the similarity matrix shows certain structures and patterns rather than being completely random. This finding suggests that even for spatially distant regions, there exists some correlation or similarity between their rainfall patterns.

From a temporal perspective, the matrix demonstrates clear short-term correlation and long-term memory effects. Each time step shows high similarity with its immediately preceding and following time steps, reflecting the continuity and smooth transition of the rainfall field on short time scales. Meanwhile, even between time steps that are far apart, a certain degree of similarity is maintained, revealing long-term dependencies in the rainfall process. This phenomenon indicates that the evolution of the rainfall field is influenced not only by short-term temporal correlations but also by long-term temporal memory effects.

Traditional convolutional neural networks have demonstrated excellent capabilities in capturing local spatial correlations. However, their limitations become increasingly apparent when faced with modeling long-range spatial associations and temporal dependencies. The local receptive field of convolutional operations restricts the network's perception and utilization of global information, leading to difficulties in achieving ideal results when processing complex spatiotemporal patterns.

In contrast, the Transformer structure, by introducing the self-attention mechanism, has shown significant advantages in capturing long-range dependencies in sequential data. The self-attention mechanism allows the network to reference all other positions in the sequence when computing the representation of each position, thereby establishing global dependencies. This mechanism breaks through the local limitations of convolutional neural networks, enabling Transformers to better understand and utilize long-range contextual information in sequential data.

However, the standard Transformer structure still has deficiencies in modeling the associations between conditional information and the sequence to be predicted. Although the self-attention mechanism can effectively capture dependencies within the sequence, it does not explicitly consider the impact of conditional information on the prediction process. This results in the model's inability to fully utilize known conditional distribution characteristics, limiting its performance in conditional sequence prediction tasks.

Based on the above analysis, we introduced a Causal Attention mechanism into the Transformer's spatiotemporal denoising model to capture the association between conditional rainfall distribution features and the rainfall sequence to be predicted. Specifically, we use the rainfall sequence to be predicted (the effect) as the Query, and the conditional rainfall latent space distribution features (the cause) as the Key-Value pairs. The attention weights between them are then calculated through Causal Attention. This design enables the model to dynamically adjust the current prediction results based on the changing trends of historical rainfall distribution, achieving adaptive prediction.

By introducing the Causal Attention mechanism, the model can better establish the causal relationship between conditional rainfall distribution features and the rainfall sequence to be predicted, thereby improving the accuracy and interpretability of the prediction.
\begin{figure*}[!t] 
  \centering	
  \includegraphics[width = \linewidth]{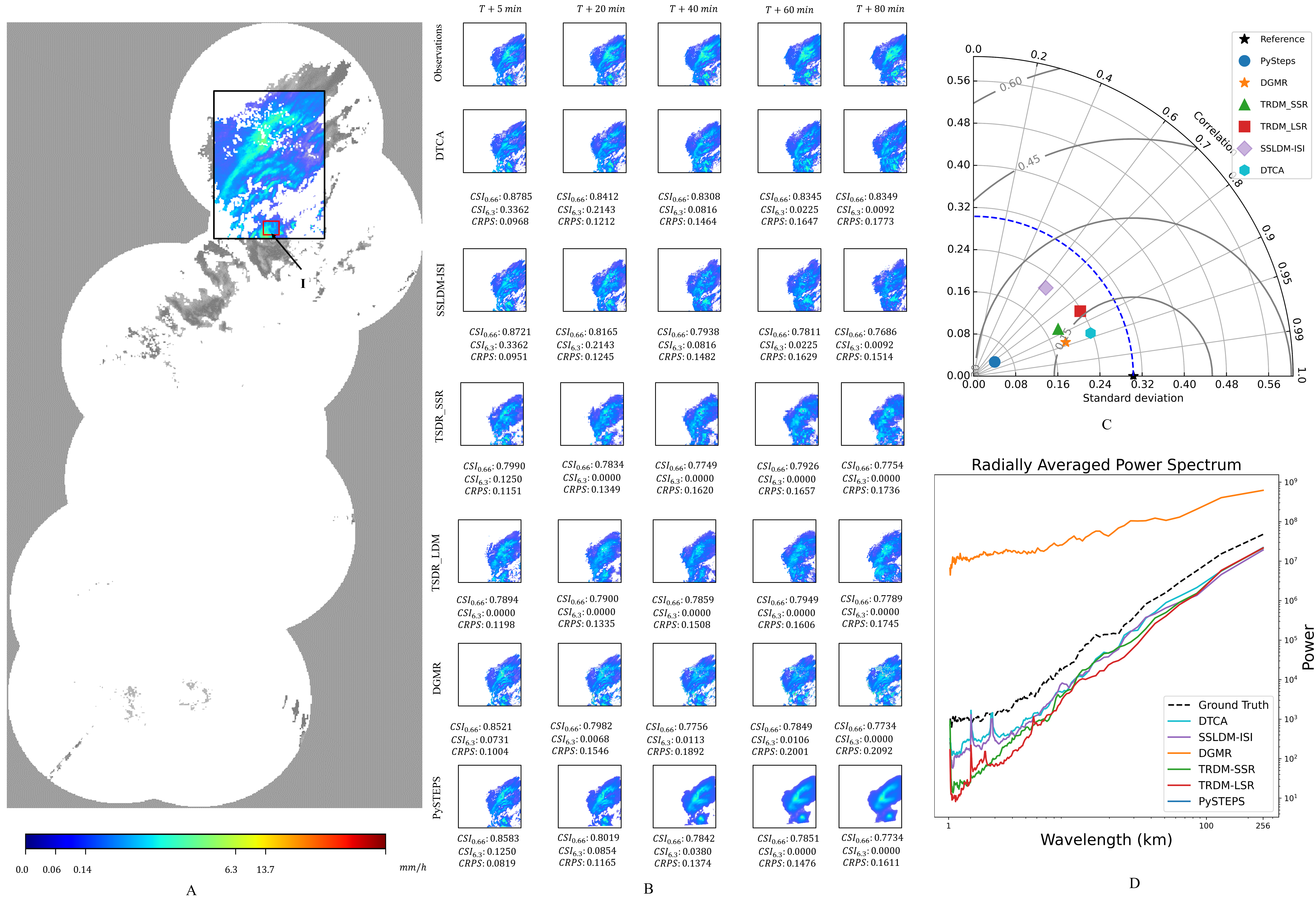}
  \caption{Prediction performance \& Visualization evaluation of Transformer Model for a rainfall event that occurred on October 07, 2021, at 04:05:00 on SW dataset.
  \textcolor{black}{A: Schematic diagram of the predicted area;
  B: Comparison of model prediction results; C: Taylor diagrams of the mean rainfall predicted by the model in Area \uppercase\expandafter{\romannumeral1}; D:Radial average power spectrum at different wavelengths (lead time 80 minutes). }}
  \label{r1}   
\end{figure*}

The process can be examined from a macro perspective, as illustrated in  \figref{f4}.C. In the proposed model, each $2\times2$ Token functions as a Query, interacting with all historical rainfall information. Through the causal attention mechanism, the current Token accesses not only rainfall information from surrounding areas but also global rainfall data. This approach enables a more comprehensive understanding of the spatial distribution and propagation patterns of rainfall. This design enables the model to establish causal relationships between the current Token and historical information, identifying temporal patterns of rainfall changes, such as periodic fluctuations or long-term trends.
Due to the introduction of the causal attention mechanism, Tokens can simultaneously capture changes in the overall weather system and small-scale local rainfall characteristics. By comparing with historical information, the model can infer causal associations between current rainfall distributions and past rainfall events, thus achieving more accurate and interpretable rainfall predictions.
The algorithm is as follows:

\begin{equation}
  \begin{gathered}
  Q = W^Q\times Z_p,K = W^K\times Z_c, V = W^V\times Z_c;\\
  \text{Flatten} =X.\text{Reshape}( 1 , \frac{(B\times C \times T \times L )}{\text{numheads} \times \text{headdim}}, \\ \text{numheads}, \text{headdim});
\end{gathered}
   \end{equation}
   
\begin{equation}
  \begin{gathered}
    Q = \text{Flatten}(Q) \\
    K = \text{Flatten}(K) \\
    V = \text{Flatten}(V)\\
    \text{Attention}i(Q_i, K_i, V_i) = \text{softmax}\left(\frac{Q_iK_i^T}{\sqrt{d{k_i}}}\right)V_i;
    \end{gathered}
  \end{equation}
  where, Q is the calculation product of the prediction result and the matrix $W^Q$, K and V are calculated by the conditions $W^K$ and $W^V$ respectively, and Di represents the length of the vector.
  To further enhance the model's understanding and representation capabilities for complex rainfall patterns, we propose the Channel-To-Batch Shift operation and combine it with the Causal Attention mechanism. The Channel-To-Batch Shift operation shifts the input data by transferring data from the channel dimension to the batch dimension. This operation enables the model to capture information correlations between different spatiotemporal points, thereby improving the coherence and consistency of predictions.
  Specifically, we reserve a portion of the sample quantity to perform the Shift operation in the batch dimension, creating a certain spatiotemporal misalignment among samples within the same batch. It's worth noting that our Channel-To-Batch Shift (CTBS) operation is implemented very concisely, requiring only three lines of code to complete (Algorithm \ref{alg:code}).
  \begin{table*}[!t]
    \centering
    \caption{The average of all metrics in the Swedish Dataset with a lead time of 80 minutes (16 frames)}
    \label{tab:AvG}
    \begin{tabularx}{\textwidth}{XXXXll}
      \hline
      Method   &CSI $0.06mm/h \uparrow $  & CSI $ 6.3mm/h\uparrow $  & FSS $\uparrow $& CSPR $\downarrow$  \\ \hline
      PySteps& 0.473&\textbf{0.0495}& 0.872&0.144&\\
      DGMR&0.433&0.0397& 0.882&0.143\\
      TRDM-SSR&0.513&0.0159&0.892&0.123\\
      TRDM-LSR&0.490&0.0144&0.891&0.143\\
      SSLDM-ISI&0.558&0.0402&0.906&0.122\\
      DTCA&\textbf{0.572}&0.0463&\textbf{0.911}&\textbf{0.115}\\
       \hline
    \end{tabularx}
    \end{table*}
    \begin{figure*}[!t]
      \centering
      \subfigure[CSI $\geqslant  0.06 mm/h$]{
      \includegraphics[width=0.22\textwidth]{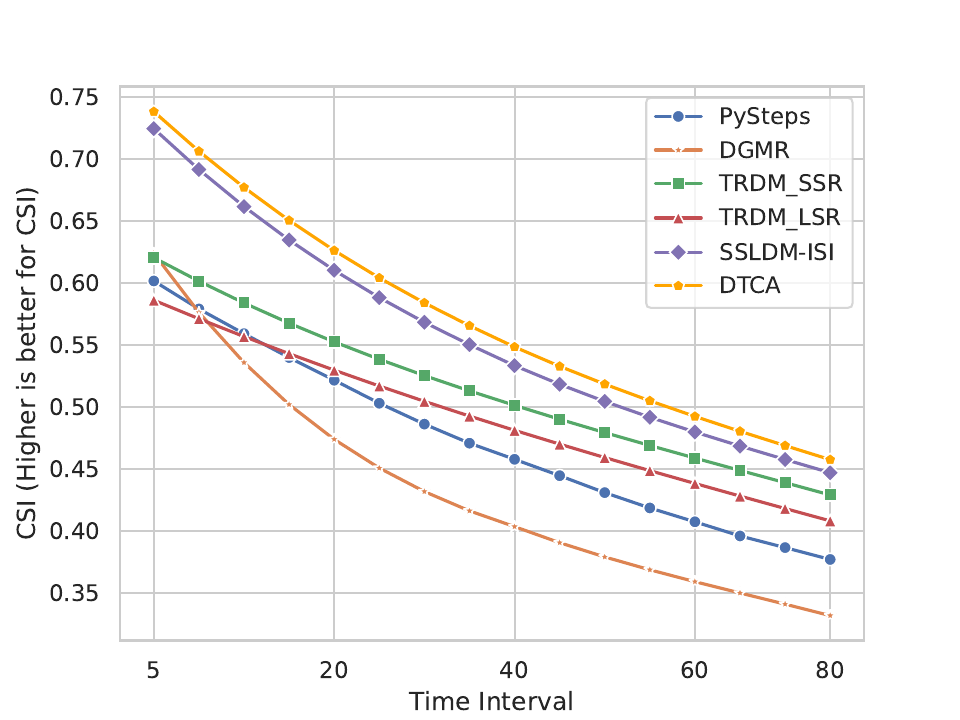}
      \label{fig:csisw}
      }
      \hfill
      \subfigure[CSI $\geqslant  6.3mm/h$]{
      \includegraphics[width=0.22\textwidth]{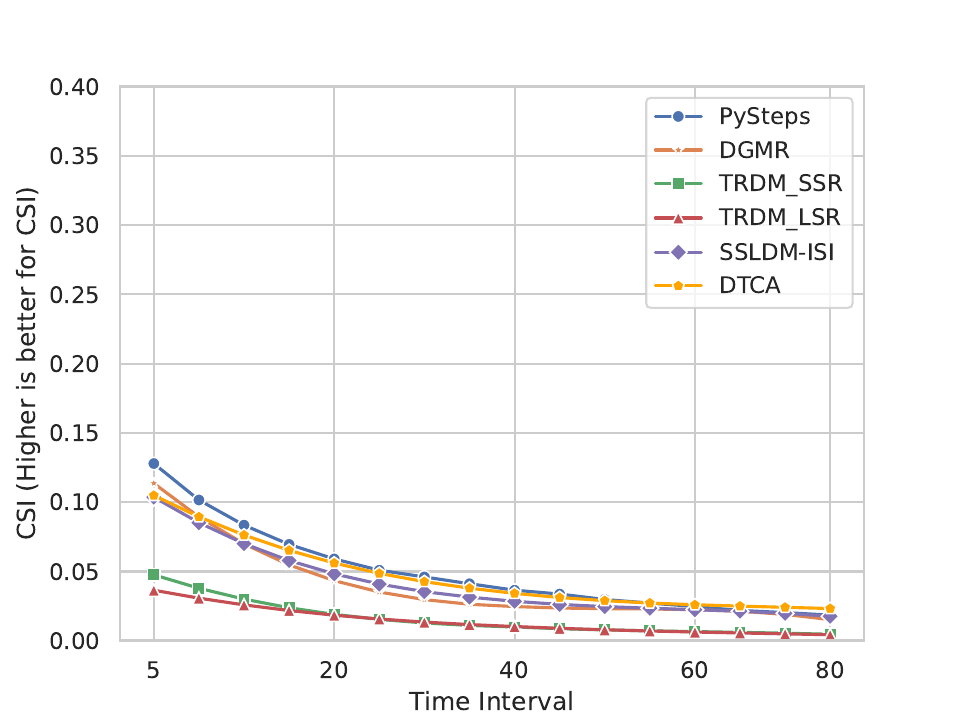}
      \label{fig:csi8sw}
      }
      \hfill
      \subfigure[CRPS]{
      \includegraphics[width=0.22\textwidth]{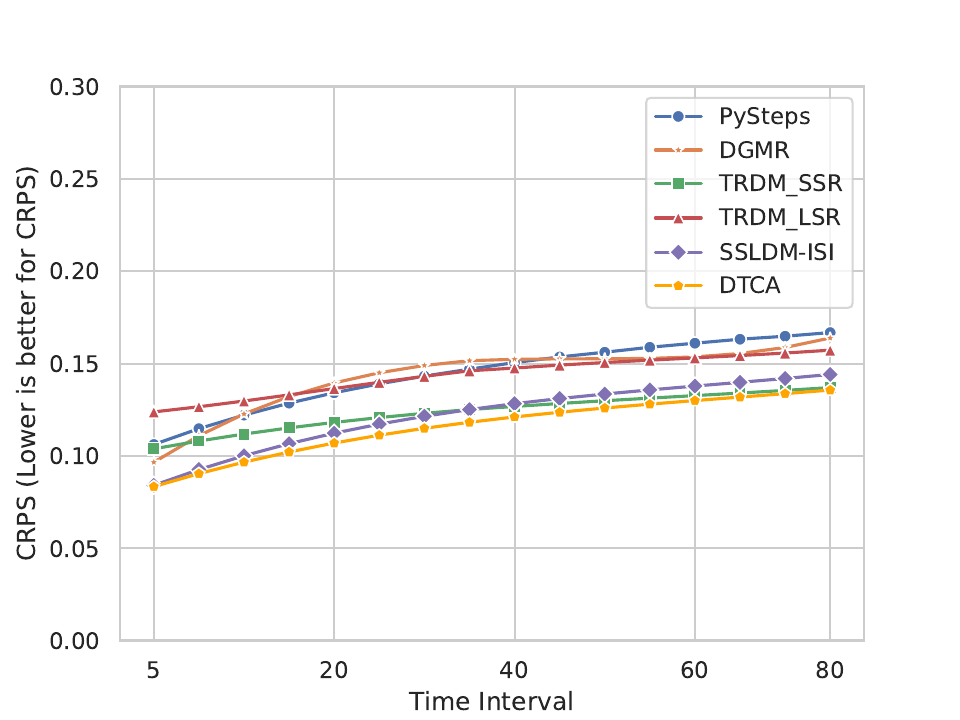}
      \label{fig:crpsw}
      }
      \hfill
      \subfigure[FSS]{
      \includegraphics[width=0.22\textwidth]{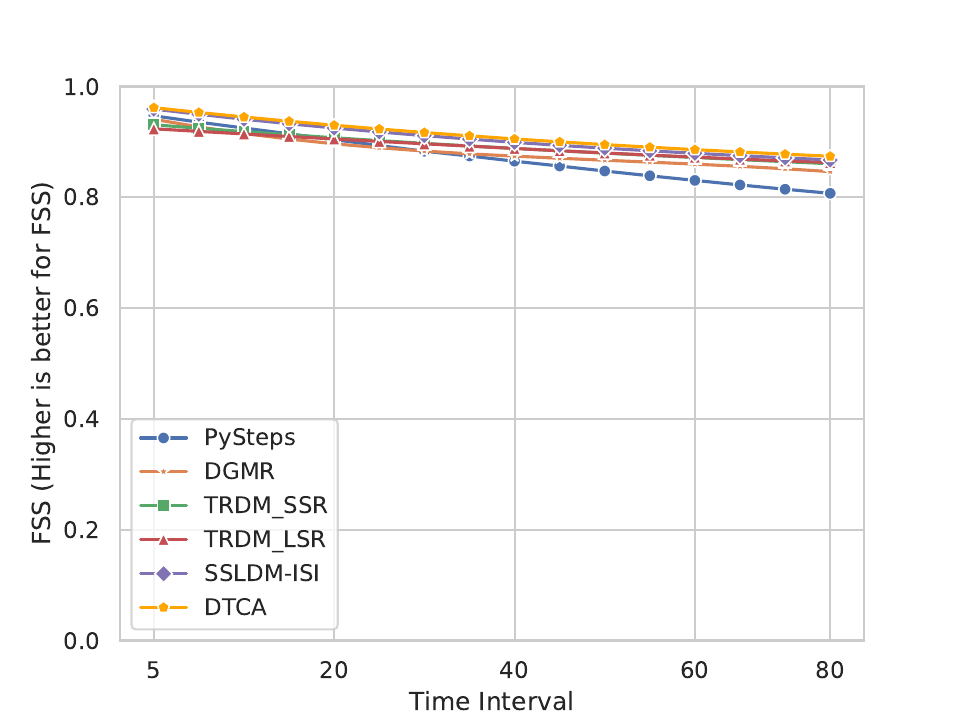}
      \label{fig:fsssw}
      }
      \hfill
      \caption{Comparative performance of Transformer and other models on SW dataset in precipitation prediction across different rainfall Intensities and prediction Times.}
      \label{fig:aszsw}
      \end{figure*}
      \begin{figure}[!t]
        \centering
        \subfigure[DTCA]{
        \includegraphics[width=0.2\textwidth]{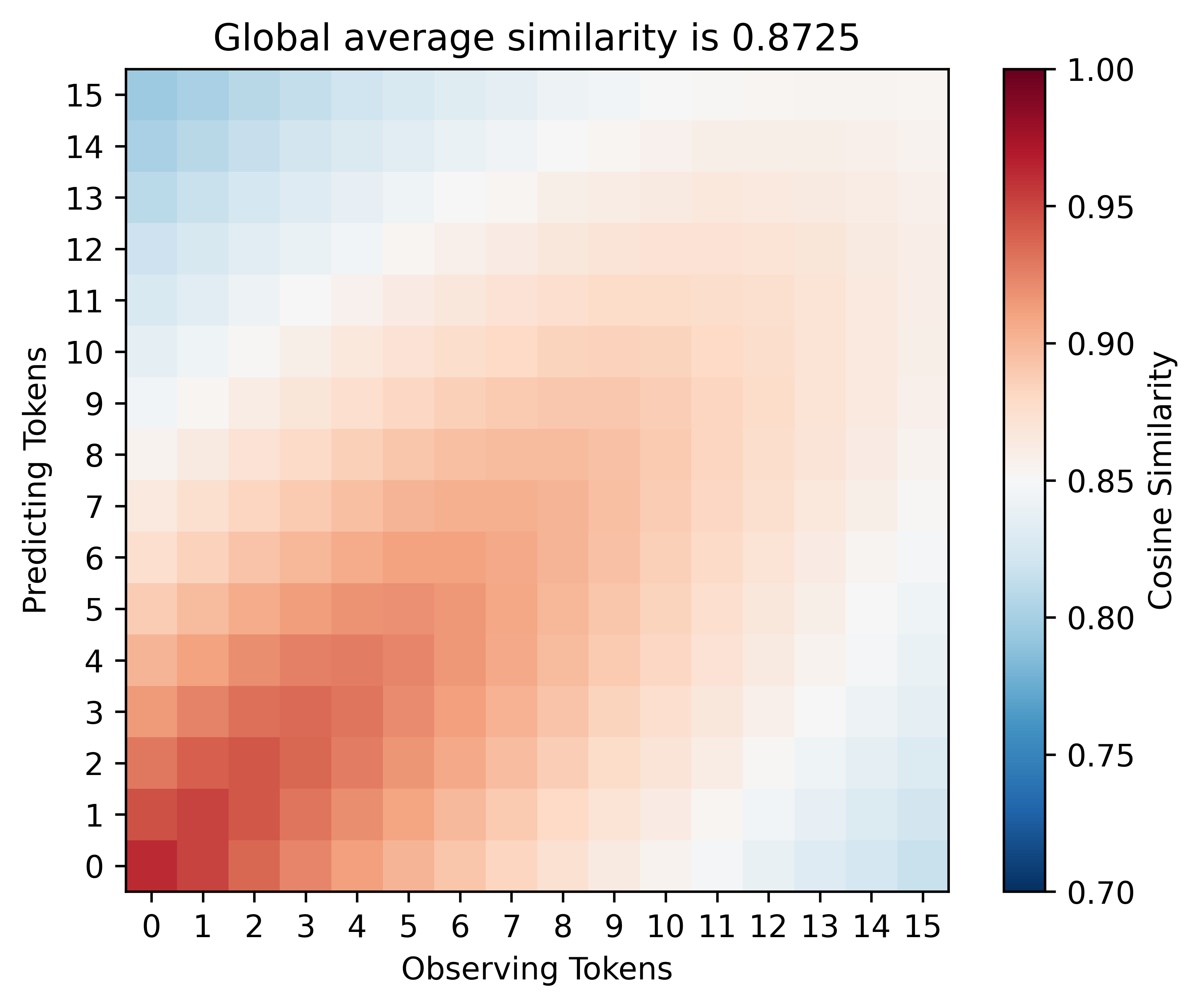}
        \label{token_1}
        }
        \hfill
        \subfigure[SSLDM-ISI]{
        \includegraphics[width=0.2\textwidth]{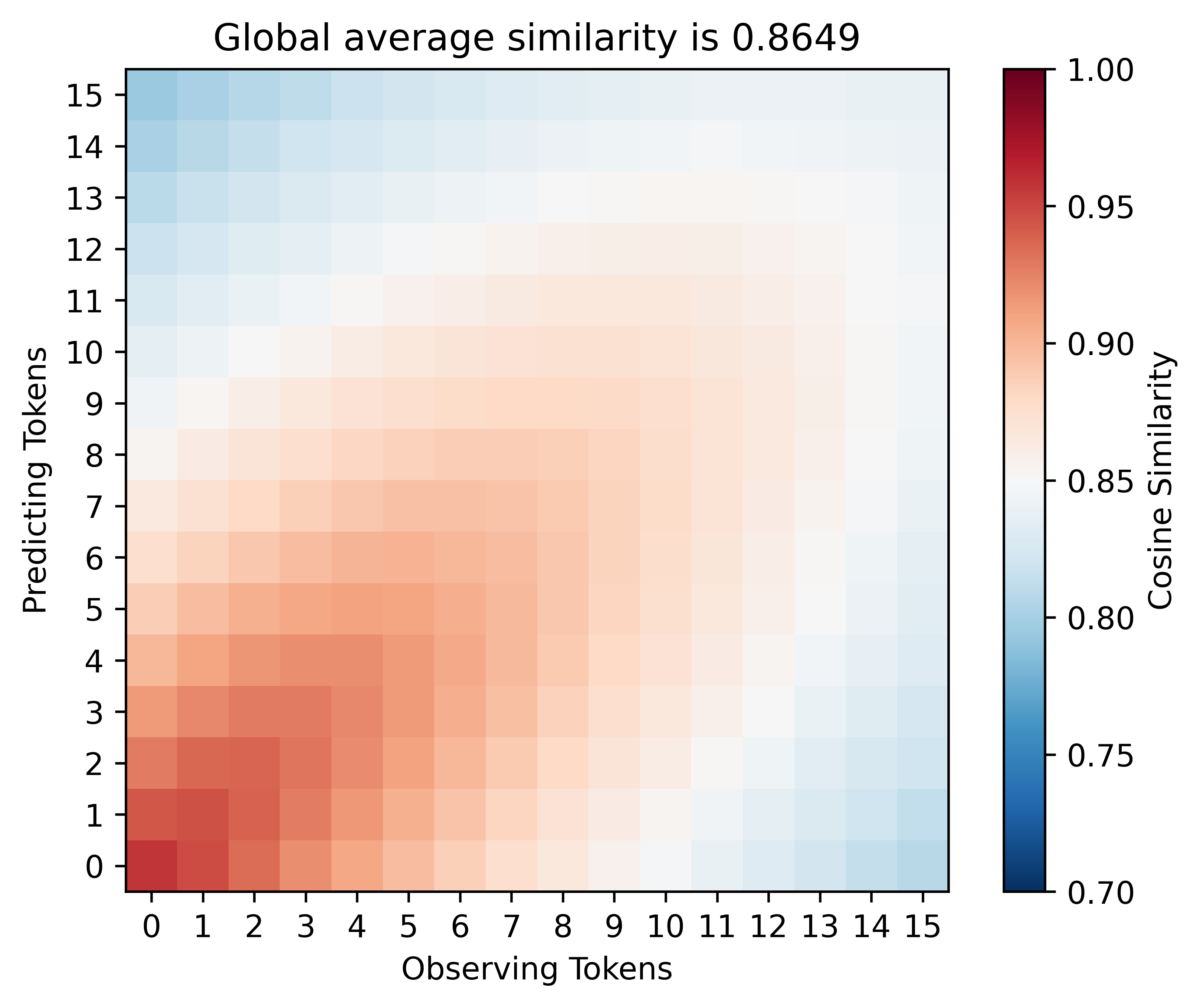}
        \label{token_2}
        }
      
        \caption{Visualization of token cosine similarity between predictions (16 frames) and observed data using the DTCA and SSLDM-ISI models on the SW Dataset.(a)DTCA,(b)SSLDM-ISI.}
        \label{token_sw}
        \end{figure}
We utilize the CTBS operation to reallocate a portion of the data from the channel dimension to the batch dimension, effectively creating a set of new "extended samples." These samples are generated by dispersing information from different channels into different batch dimensions, with each extended sample containing different parts of the original Token information. This operation aims to increase data diversity by expanding channel information to the batch dimension, thereby enhancing feature richness and model accuracy.

To ensure that each extended sample can effectively capture meaningful information, we introduce a Global Contribution Linear Layer. This layer maps the features of each extended sample to a unified feature space and extracts and aggregates these features through linear transformation. In this way, each extended sample can independently learn key feature patterns without relying on assistance from other samples or channels. At the same time, it also avoids the problem of gradient disappearance caused by introducing additional conditions.
When computing multi-head cross-attention, we reorganize the dimensions $(B, (C, F), N)$ into $(1, dim, headnum, headdim)$, where headnum is the number of heads in multi-head attention, and headdim is the dimension of each head. Through this reorganization, we flatten the batch dimension B and channel dimension $C \times F$, allowing each virtual sample to interact independently with other samples during attention computation.
When the reorganized tensor is used as the query matrix Q for dot product calculation with key matrix K and value matrix V, there will be no confusion in the correspondence of batch dimensions since the batch dimension has already been flattened.

\section{Experment}
\subsection{Dataset}
We conducted comprehensive experimental evaluations on the Sweden dataset and the MSMR dataset to validate the superiority and broad applicability of our proposed method. These two datasets can be accessed through the following links.

\subsection{Other Methods and Verification Scores}
To evaluate the advantages of our model, we compared it with several state-of-the-art methods in rainfall prediction, including Pysteps (based on statistical learning), DGMR (based on GAN), and diffusion model-based approaches such as TRDM and SSLDM-ISI. For comprehensive assessment, we employed a range of widely used evaluation metrics in rainfall prediction, including CSI, CRPS \cite{gneiting2007strictly}, and FSS \cite{skillful}. These metrics allow us to thoroughly assess different aspects of prediction performance, from overall accuracy to spatial structure preservation and probabilistic forecast skill.

\begin{figure*}[!t] 
  \centering	
  \includegraphics[width = \linewidth]{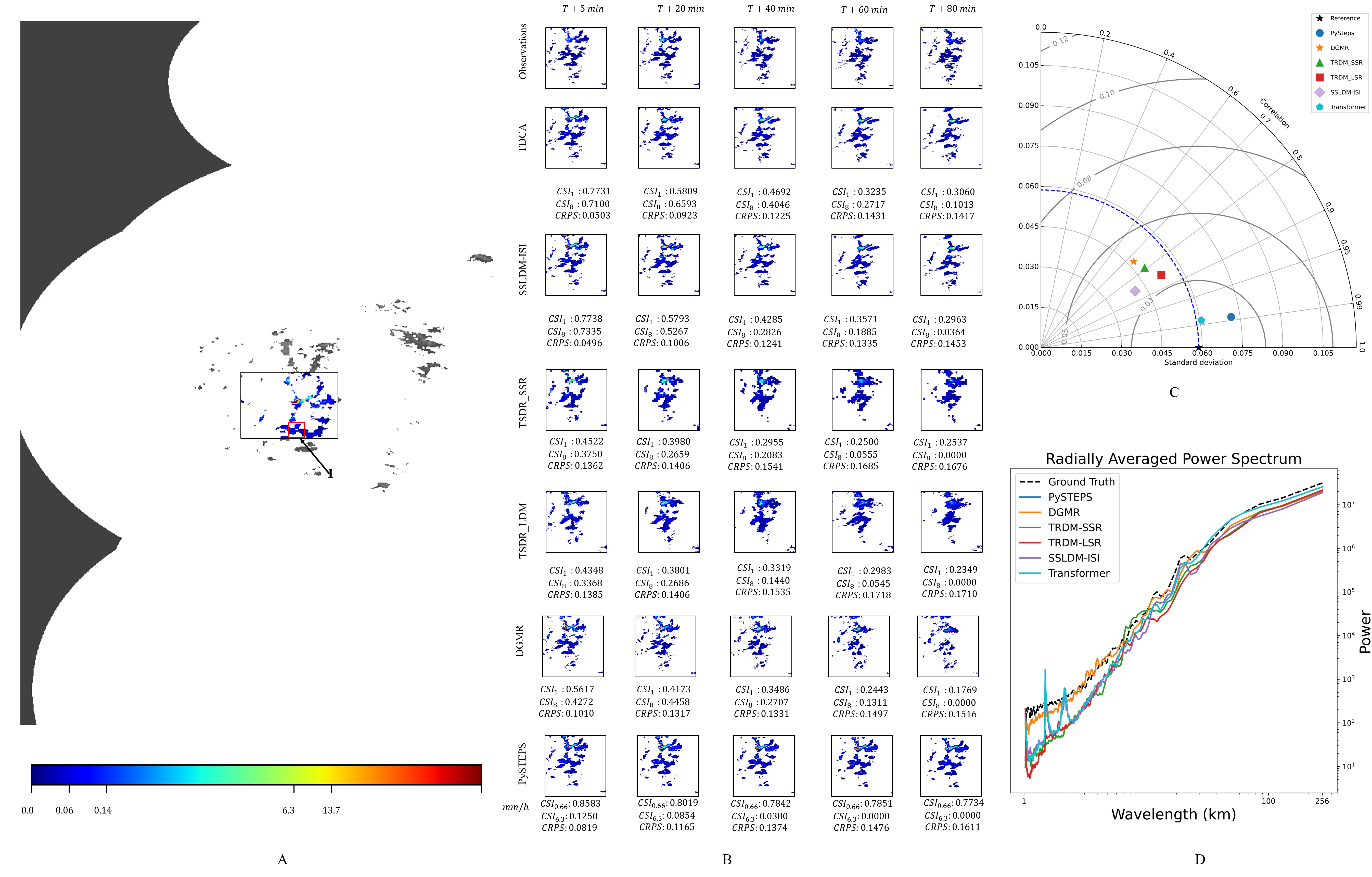}
  \caption{Prediction performance \& Visualization evaluation of Transformer Model for a rainfall event that occurred on April 09, 2022, at 13:56:00 on MRMS dataset.
  \textcolor{black}{A: Schematic diagram of the predicted area;
  B: Comparison of model prediction results; C: Taylor diagrams of the mean rainfall predicted by the model in Area \uppercase\expandafter{\romannumeral1}; D:Radial average power spectrum at different wavelengths (lead time 80 minutes)}}
  \label{r2}
\end{figure*}
\begin{table*}[!t]
  \centering
  \caption{The average of all metrics in the MRMS dataset with a lead time of 80 minutes (16 frames)}
  \label{tab:AvGmrms}
  \begin{tabularx}{\textwidth}{XXXXll}
    \hline
    Method   &CSI $1 mm/h \uparrow $  & CSI $ 8 mm/h\uparrow $  & FSS $\uparrow $& CRPS $\downarrow$  \\ \hline
    PySteps& 0.3383&0.0249& 0.5793&0.1675\\
    DGMR&0.3676&0.0225&0.6330&0.1542\\
    TRDM-SSR&0.3560&0.0125&0.6146&0.1542\\
    TRDM-LSR&0.3380&0.0144&0.6131&0.1553\\
    SSLDM-ISI &0.4161&0.0247&0.7044&0.1351\\
    DTCA&\textbf{0.4235}&\textbf{0.0267}&\textbf{0.7117}&\textbf{0.1310}\\
     \hline
  \end{tabularx}
  \end{table*}

\subsection{Results on Swedish}

The \figref{r1} shows prediction examples from various models on the MRMS dataset. We present forecast images generated by each method for a $256km \times 256km$ area at time points $T+5$, $T+20$, $T+40$, $T+60$, and $T+80$ minutes. We employed CSI (with thresholds of $0.06mm/h$ and $6.3 mm/h$) and CRPS as evaluation metrics. The assessment results are displayed below each set of prediction images, along with Taylor diagrams and radially averaged power spectra. These comprehensive evaluations demonstrate that the DTCA model excels across multiple performance indicators.

\tableref{tab:AvG} and \figref{fig:aszsw} compare the performance of six precipitation prediction methods on the Swedish dataset, with a forecast lead time of 80 minutes. We used three metrics - CSI, FSS, and CRPS - to evaluate the prediction effectiveness of each method. The CSI metric was calculated at two thresholds, 0.06mm/h and 6.3mm/h, to assess the models' prediction capabilities at different precipitation intensities.
The results show that the DTCA method achieved the best performance in CSI (0.06mm/h), FSS, and CRPS, with scores of 0.572, 0.911, and 0.115 respectively, demonstrating its advantages in predicting light rainfall, overall similarity, and probabilistic forecast skill. For the CSI (6.3mm/h) metric, DTCA performed similarly to the comparable PySteps method, with scores of 0.0463 and 0.0495 respectively. However, DTCA's ability to predict heavy rainfall scenarios surpassed that of other deep learning methods.

\subsubsection{ Token Similarity Analysis }
We analyzed the cosine similarity of Tokens in the latent space during the prediction process between two models: DTCA and SSLDM-ISI. As shown in \figref{token_sw}, both models demonstrate high similarity with the observed results in the first few frames, but this similarity gradually decreases as the prediction time increases. Notably, after the 13th frame, the Unet-based SSLDM-ISI model shows a significant decrease in similarity with the observed results, while the DTCA model maintains a higher correlation even after the 13th frame.

This indicates that although both models can effectively capture the characteristics of the observed results in the early frames, the DTCA model exhibits stronger stability and higher robustness in long-term predictions, maintaining high similarity with the observed results for a longer period. The global average similarity of DTCA is 0.8725, higher than SSLDM-ISI's 0.8649. This further validates the advantages of the DTCA model in processing time-series data, especially in prediction tasks requiring longer time spans. Its structure and mechanism make it more adaptive and accurate compared to the SSLDM-ISI model.

    \subsection{Results on MRMS}

\begin{figure}[!t]
            \centering
            \subfigure[CSI $\geqslant 1 mm/h$]{
            \includegraphics[width=0.2\textwidth]{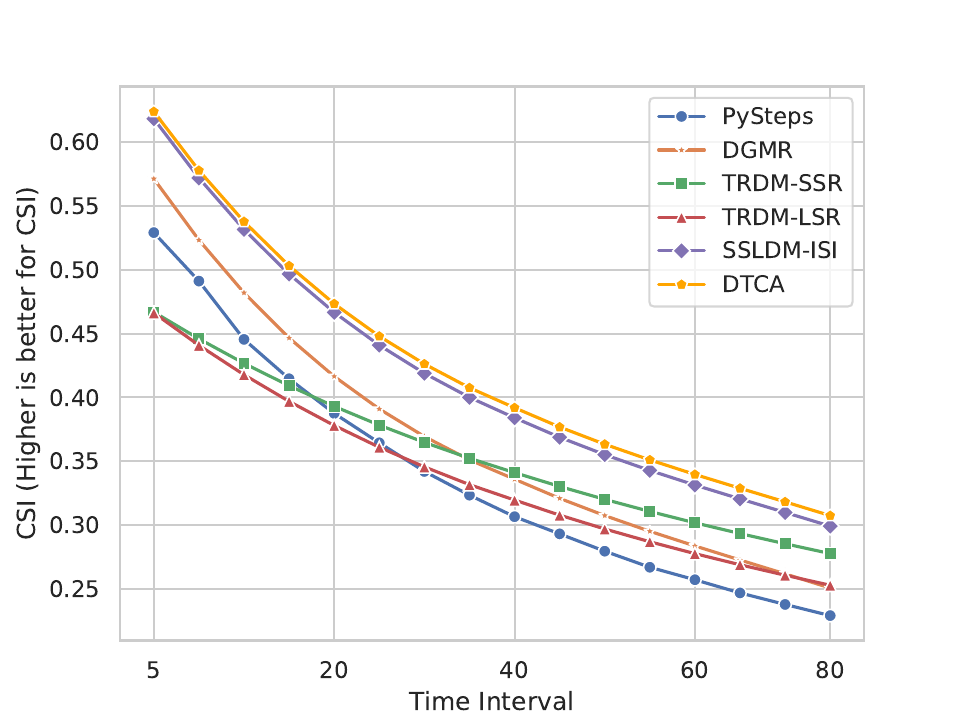}
            \label{fig:csimrms}
            }
            \hfill
            \subfigure[CSI $\geqslant 8 mm/h$]{
            \includegraphics[width=0.2\textwidth]{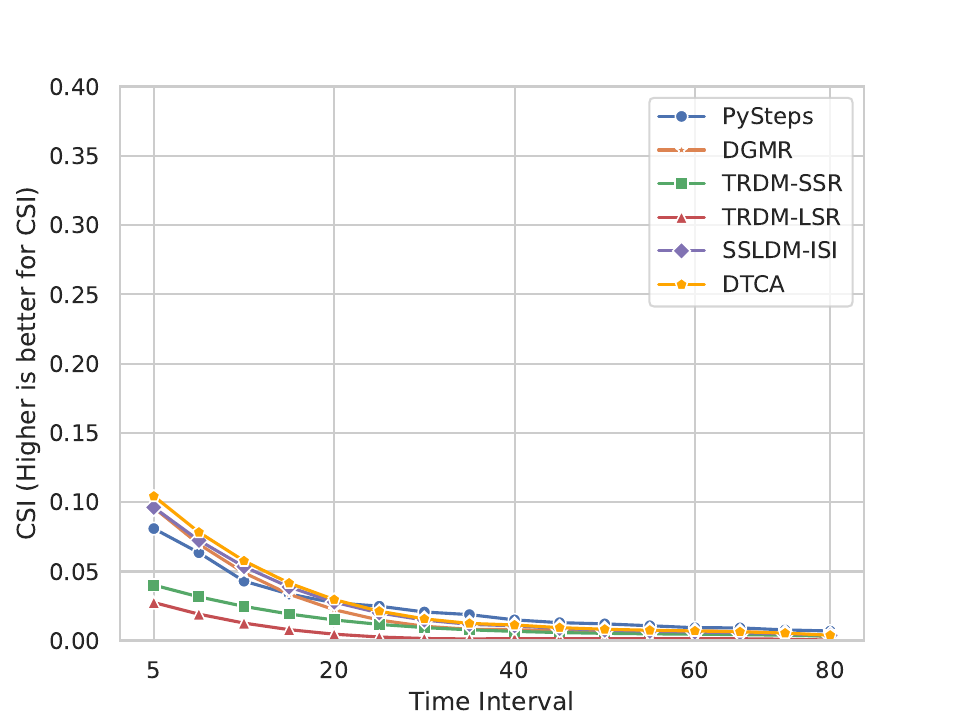}
            \label{fig:csi8mrms}
            }
            \subfigure[CRPS]{
            \includegraphics[width=0.2\textwidth]{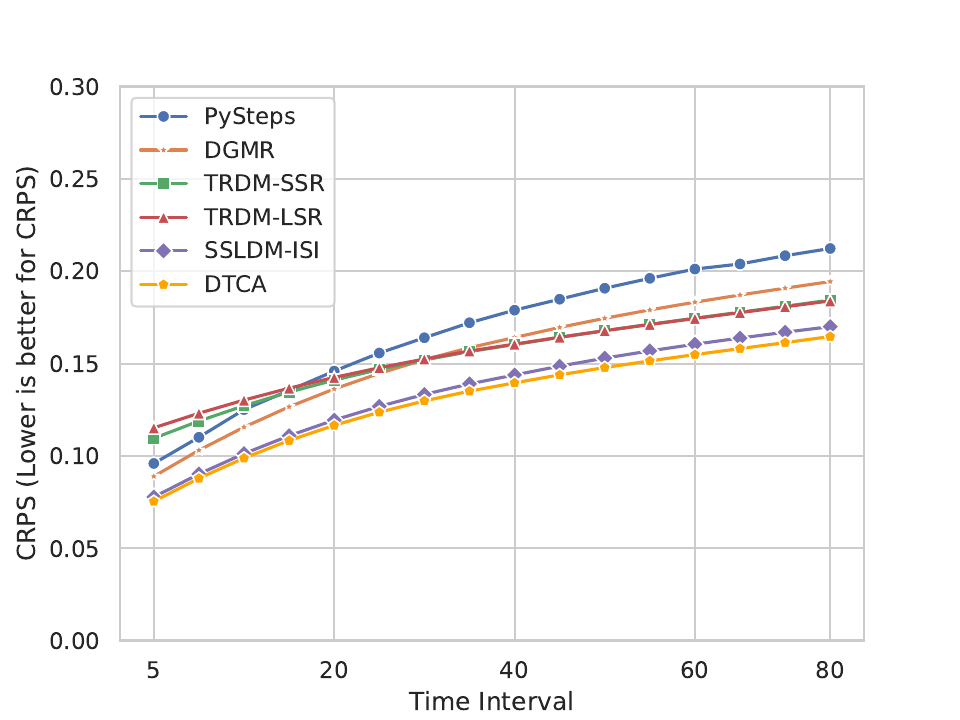}
            \label{fig:crpsmrms}
            }
            \hfill
            \subfigure[FSS]{
            \includegraphics[width=0.2\textwidth]{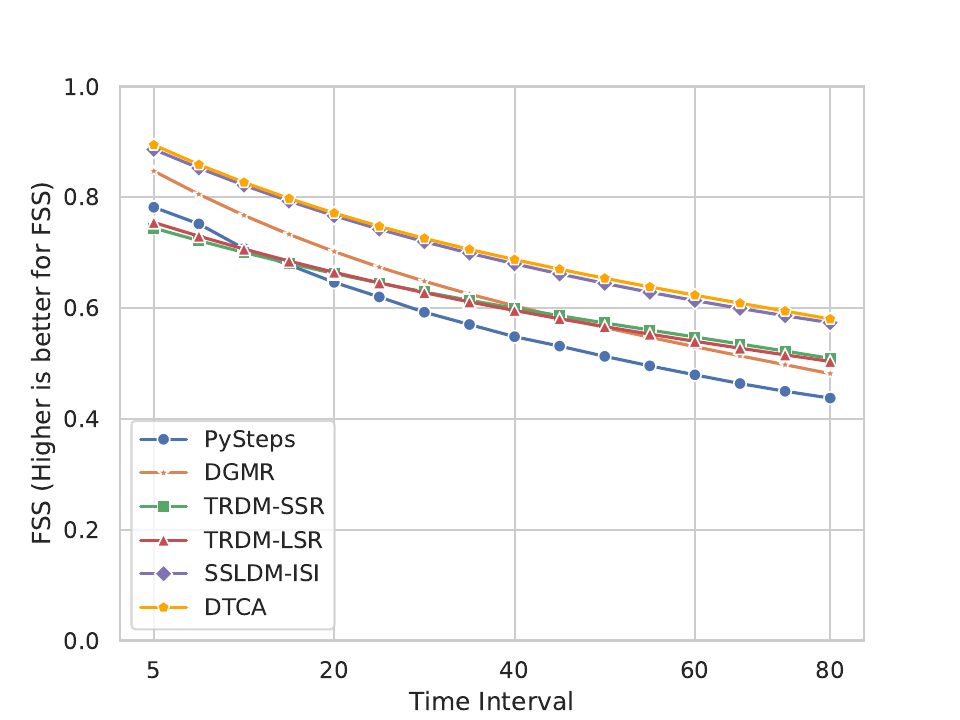}
            \label{fig:fssmrms}
            }
            \caption{Comparative performance of Transformer and other models on MRMS dataset in precipitation prediction across different rainfall Intensities and prediction Times.}
            \label{fig:aszmrms}
            \end{figure}

          In experiments using the MRMS dataset, we compared the performance of various models over an 80-minute prediction span (16 frames) (see \tableref{tab:AvGmrms} and \figref{fig:aszmrms}). We used the CSI metric to evaluate model performance at two rainfall rates: 1 mm/h and 8 mm/h.
          At the 1 mm/h rainfall rate, the DTCA model maintained high prediction accuracy throughout the entire prediction period, achieving a total score of 0.4235, which is approximately 1.7\% higher than SSLDM-ISI. In the heavy rainfall scenario (8 mm/h), DTCA scored 0.0267, significantly outperforming Pysteps (0.0249) and SSLDM-ISI (0.0247) by 7\% and 8\% respectively.
          For the CRPS metric, DTCA maintained the lowest score at all prediction time    points, indicating that its predicted probability distribution aligns most closely with the actual rainfall distribution. Similarly, for the FSS metric, DTCA maintained the highest score at all time points, surpassing the best-performing SSLDM-ISI model by approximately 2.9\% and 1\% respectively, highlighting its excellent ability to capture spatial distribution characteristics of rainfall.

            \begin{figure}[!h]
              \centering
              \subfigure[Transformer]{
              \includegraphics[width=0.2\textwidth]{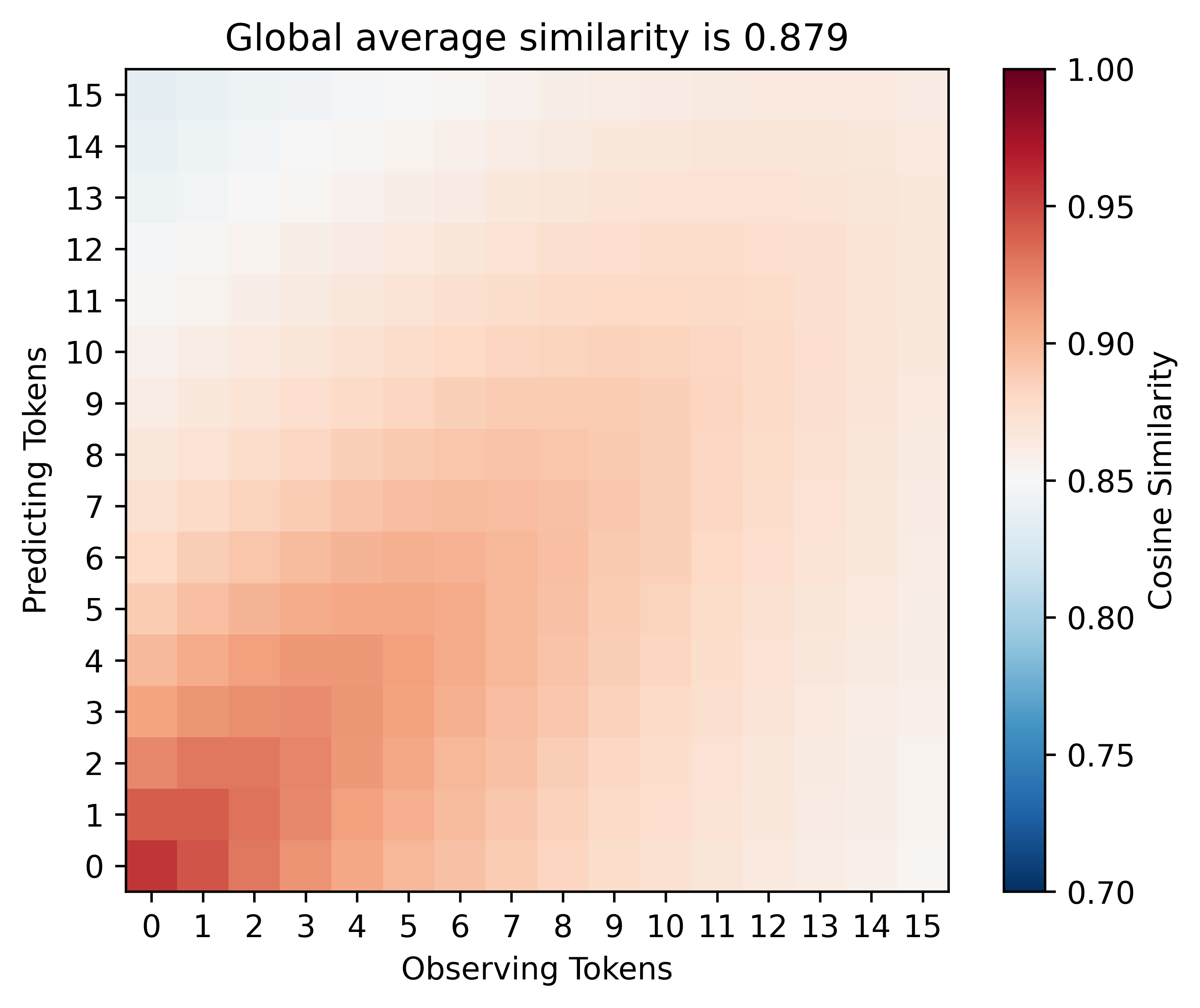}
              \label{token_3}
              }
              \hfill
              \subfigure[SSLDM-ISI]{
              \includegraphics[width=0.2\textwidth]{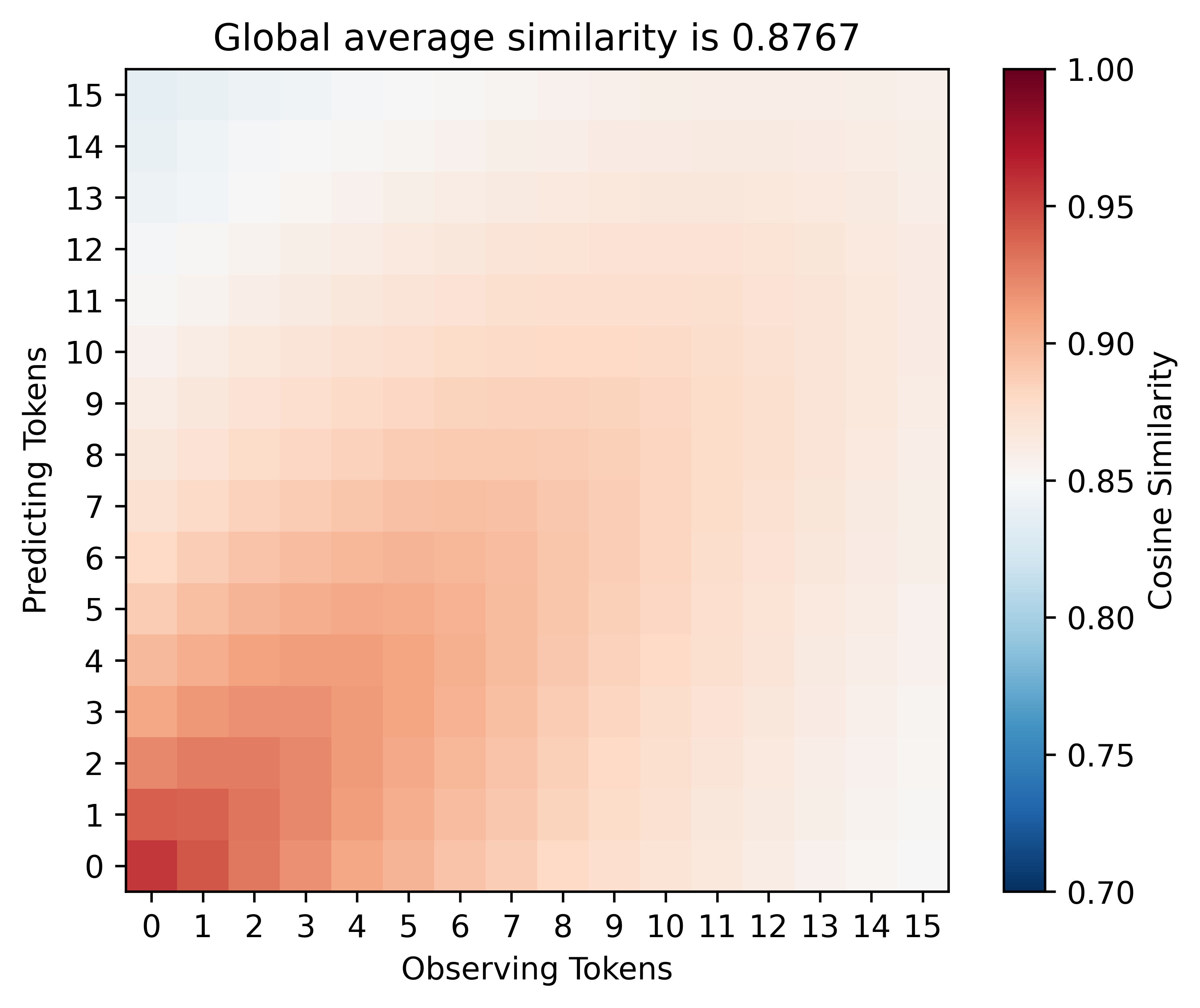}
              \label{token_4}
              }
            
              \caption{Visualization of token cosine similarity between predictions (16 frames) and observed data using the Transformer and SSLDM-ISI models on the MRMS Dataset.(a)Transformer,(b)SSLDM-ISI.}
              \label{token_mrms}
              \end{figure}

\subsubsection{Token Similarity Analysis on MRMS Dataset}
Following the method described in Section \ref{cacts32}, we converted the prediction results of DTCA and SSLDM-ISI models into Tokens and calculated their cosine similarity. As shown in \figref{token_mrms}, DTCA achieves a global average similarity of 0.879, while SSLDM's global similarity is 0.8767. This indicates that DTCA has a certain advantage in capturing global features, which translates into a leading position in time series prediction.
 In \figref{fig:aszmrms}, it can be clearly observed that the performance of all models in the rainfall prediction task gradually declines over time. However, DTCA's performance decreases at a slower rate, further demonstrating its stability and robustness in handling long time series and complex temporal patterns.

 \section{Ablation experiment}
\subsection{Space-time capture structure}

\begin{table*}[!t]
  \centering
  \caption{Average of all metrics for the 4 variants on the Swedish dataset with a lead time of 80 minutes (16 frames).}
  \label{tab:ab}
  \begin{tabularx}{\textwidth}{XXXXll}
    \hline
    Model Name   &CSI$\geqslant  0.06mm/h \uparrow $  & CSI$\geqslant 6.3mm/h\uparrow $  & FSS $\uparrow $& CSPR $\downarrow$ &\\ \hline
    DTCA$_{FST}$& \textbf{0.5724}&0.0457&\textbf{0.9110}&\textbf{0.116}\\
    DTCA$_{S+T}$&0.5660&\textbf{0.0475}& 0.9094&0.1197\\
    DTCA$_{HST+S}$&0.5700&0.0466&0.9090&0.1180\\
    DTCA$_{HST+T}$&0.5646&0.0461&0.9070&0.1200\\
     \hline      
  \end{tabularx}
  \end{table*}   
  \begin{table*}[!t]
    \centering
    \caption{Average of all metrics for the 4 variants on the Swedish dataset with a lead time of 80 minutes (16 frames)} 
    \label{tab:ab1}
    \begin{tabularx}{\textwidth}{XXXXll}
      \hline
      Model Name   &CSI$\geqslant  0.06mm/h \uparrow $  & CSI$\geqslant 6.3mm/h\uparrow $  & FSS $\uparrow $& CSPR $\downarrow$ &\\ \hline
      No Causal Attention& \textbf{0.5724}&0.0457&0.911&0.1161\\
      Causal Attention&0.5713&0.0441& \textbf{0.912}&\textbf{0.1154}\\
      Causal Attention 4&\textbf{0.5724}&\textbf{0.0463}&\textbf{0.912}&0.1158\\
      Causal Attention 8&0.5714&0.0449&\textbf{0.912}&0.1157\\
      Causal Attention 16&0.5715&0.0461&\textbf{0.912}&0.1156\\
       \hline
    \end{tabularx}
    \end{table*}

    \begin{table*}[!t]
      \centering
      \caption{Average of all metrics for the 4 variants on the MRMS dataset with a lead time of 80 minutes (16 frames)}
      \label{tab:ab2}
      \begin{tabularx}{\textwidth}{XXXXll}
        \hline
        Model Name   &CSI$\geqslant  1mm/h \uparrow $  & CSI$\geqslant 8mm/h\uparrow $  & FSS $\uparrow $& CSPR $\downarrow$ &\\ \hline
        No Causal Attention& 0.4052&0.0260&0.6832&0.1454\\
        Causal Attention&0.4151&0.0264& 0.6970&0.1390\\
        Causal Attention 4&0.4236&0.0265& \textbf{0.7117}&\textbf{0.1310}\\
        Causal Attention 8&0.4187&\textbf{0.0266}&0.6984&0.1383\\
        Causal Attention 16&\textbf{0.4238}&0.0259&0.7108&0.1316\\
      
         \hline
      \end{tabularx}
      \end{table*}

      \begin{figure*}[!t]
        \centering
        \subfigure[CSI $\geqslant  0.06 mm/h$]{
        \includegraphics[width=0.22\textwidth]{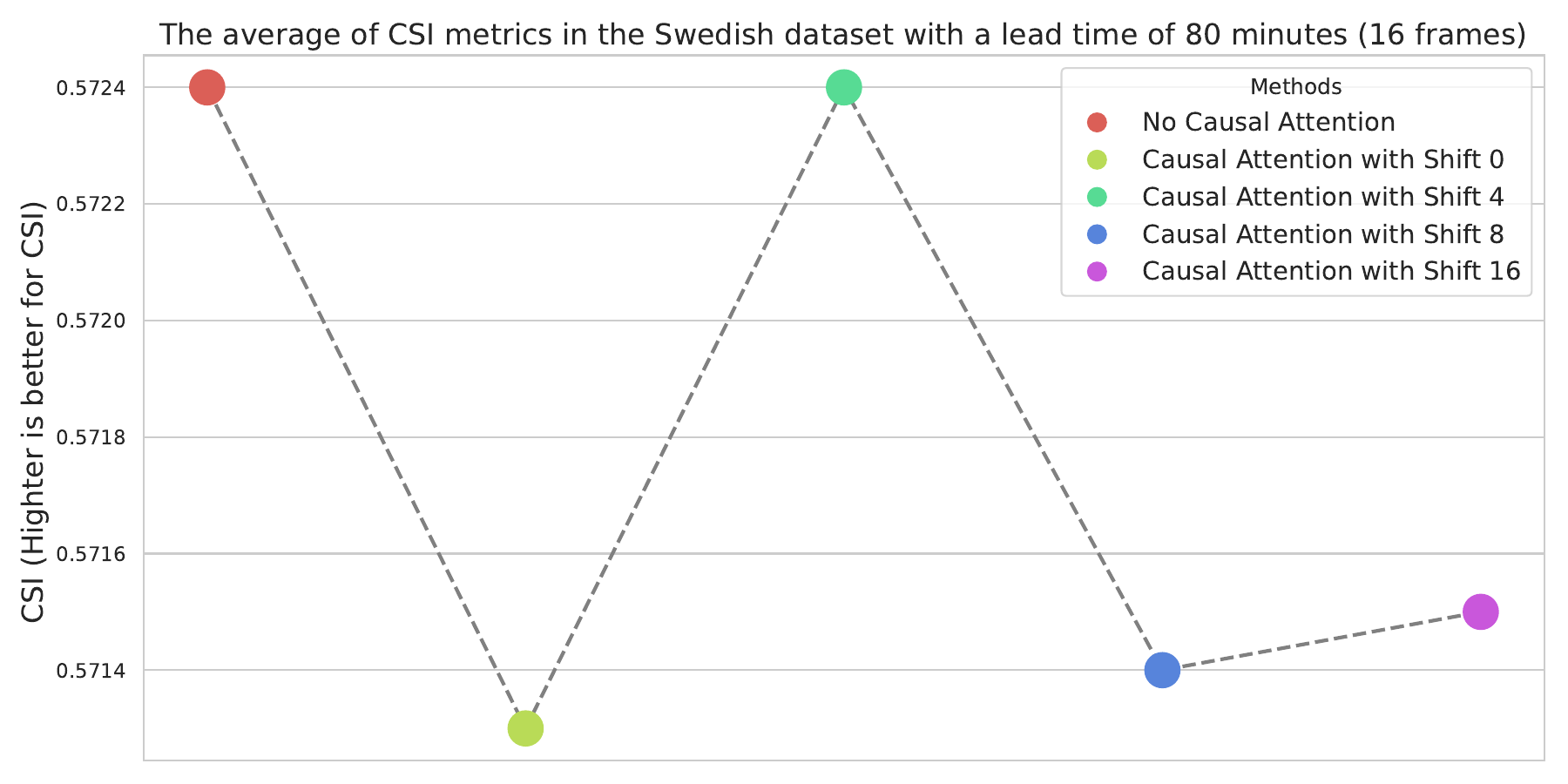}
        \label{fig:csi}
        }
        \hfill
        \subfigure[CSI $\geqslant  6.3mm/h$]{
        \includegraphics[width=0.22\textwidth]{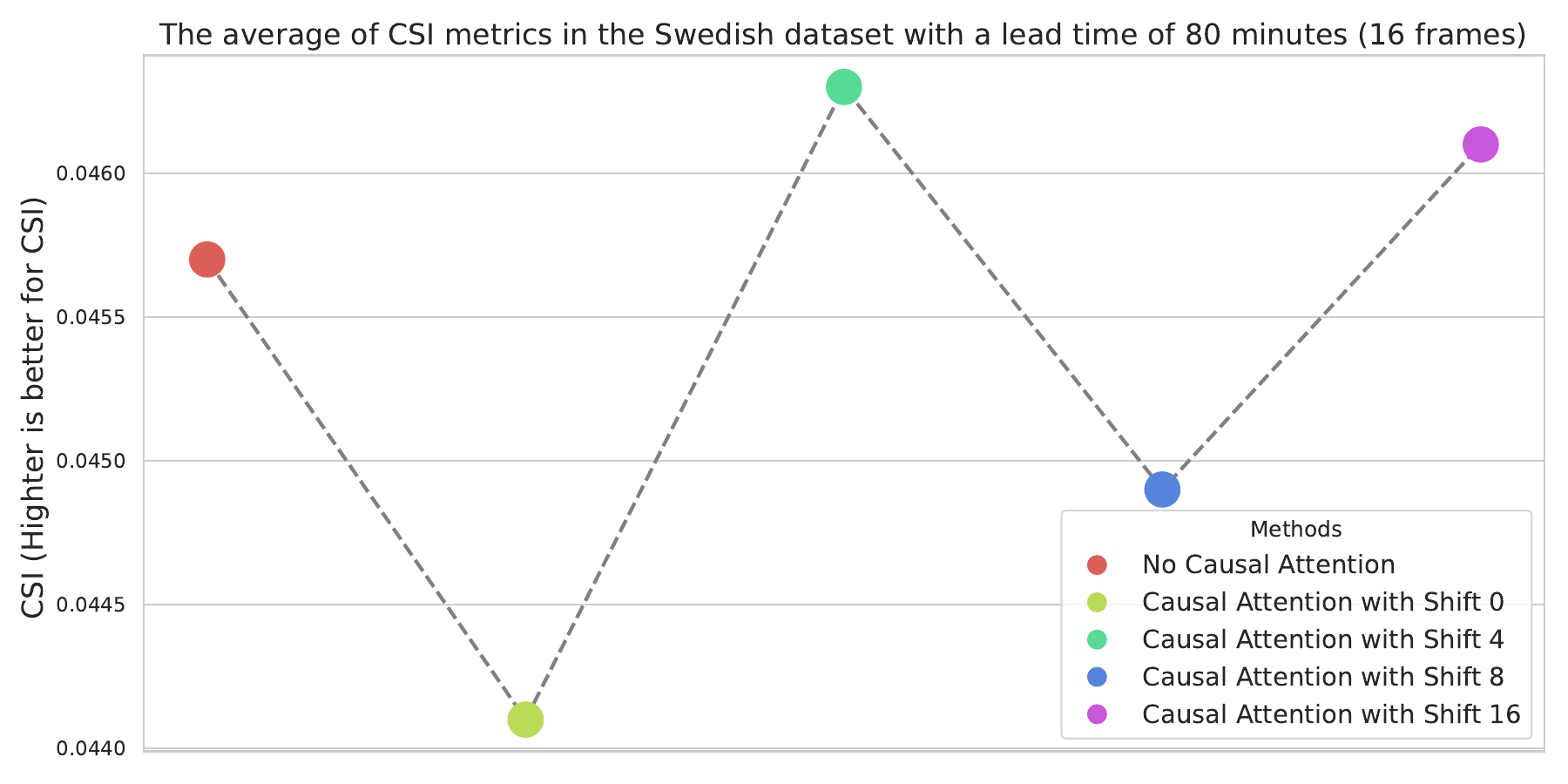}
        \label{fig:csi35}
        }
        \subfigure[CRPS]{
        \includegraphics[width=0.22\textwidth]{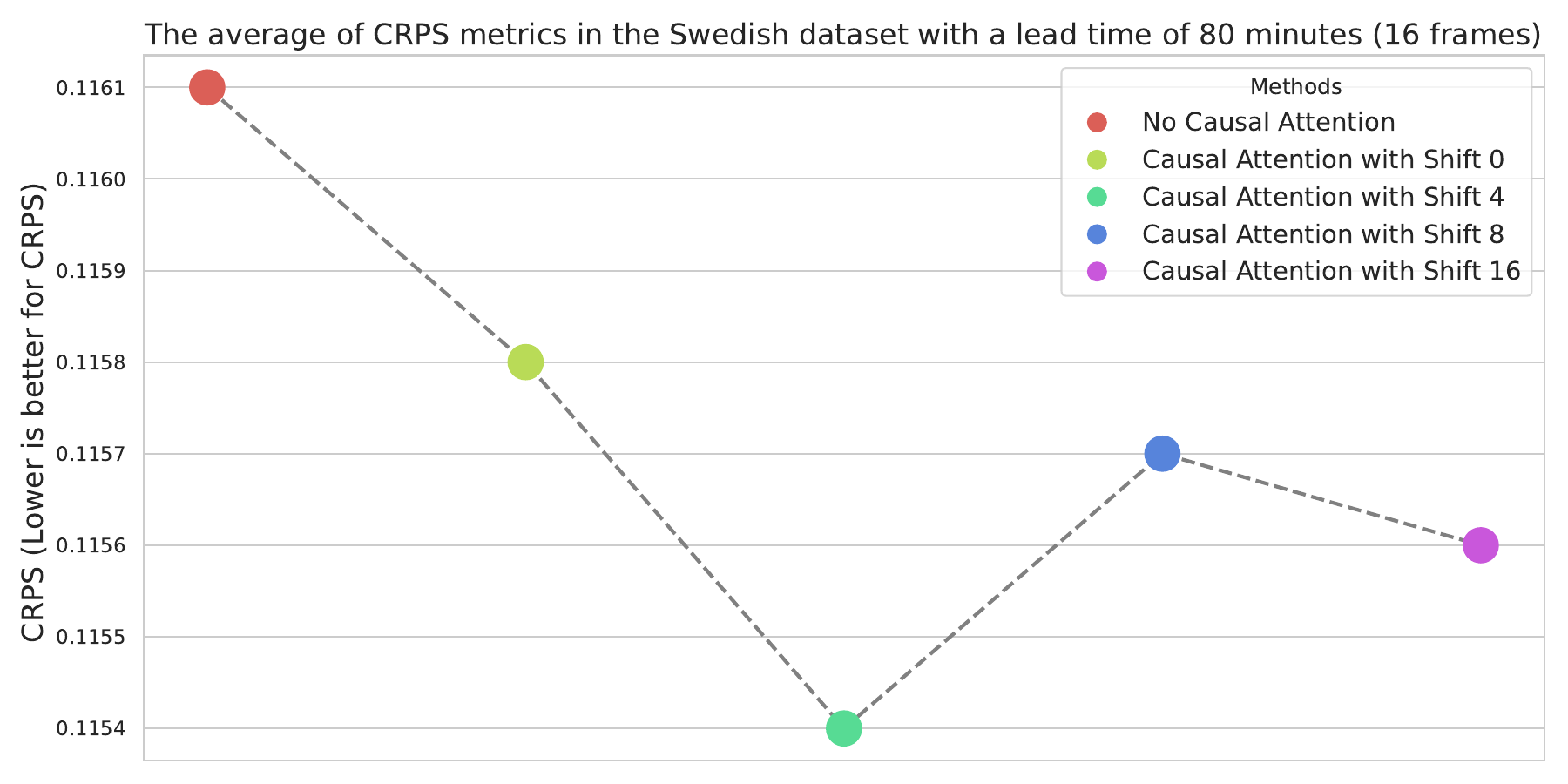}
        \label{fig:crps}
        }
        \hfill
        \subfigure[FSS]{
        \includegraphics[width=0.22\textwidth]{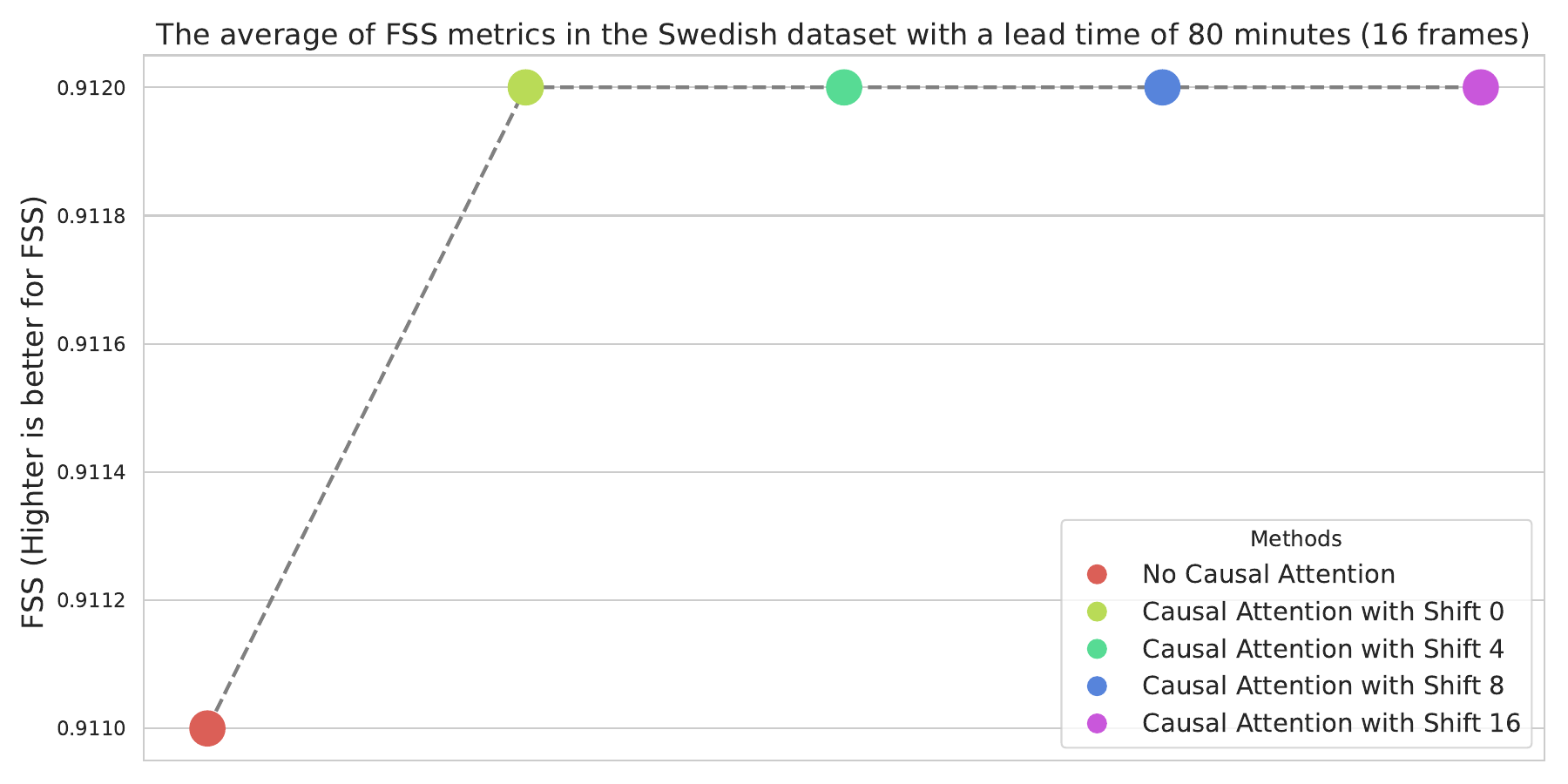}
        \label{fig:fss}
        }
        \caption{Comparison of precipitation prediction performance for 80 minutes on the Swedish dataset, with and without Causal Attention and different numbers of Channel-To-Batch Shifts.}
        \label{fig:asz}
        \end{figure*}
        \begin{figure*}[!t]
          \centering
          \subfigure[CSI $\geqslant  1 mm/h$]{
          \includegraphics[width=0.22\textwidth]{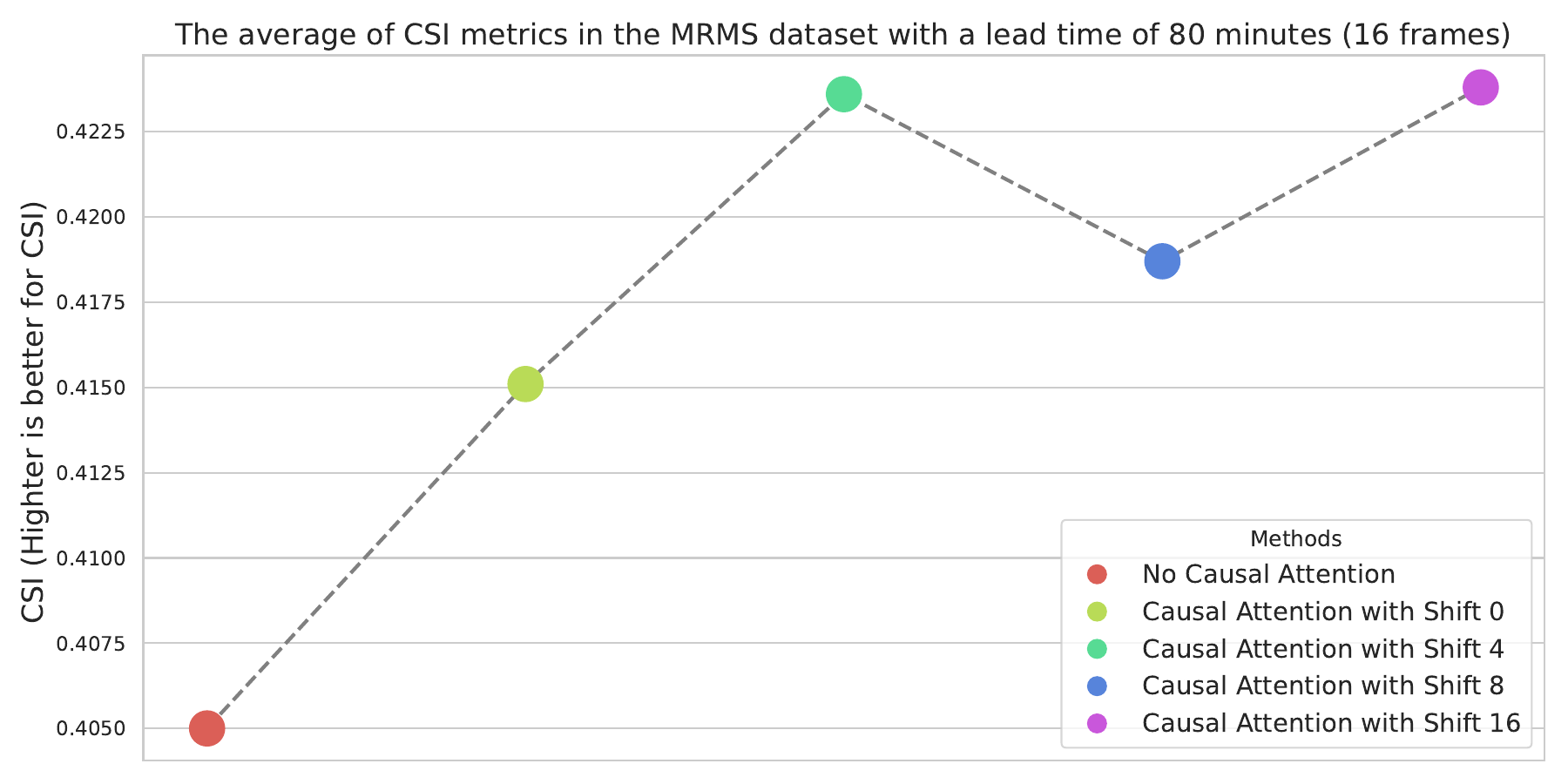}
          \label{fig:csi1}
          }
          \hfill
          \subfigure[CSI $\geqslant  8mm/h$]{
          \includegraphics[width=0.22\textwidth]{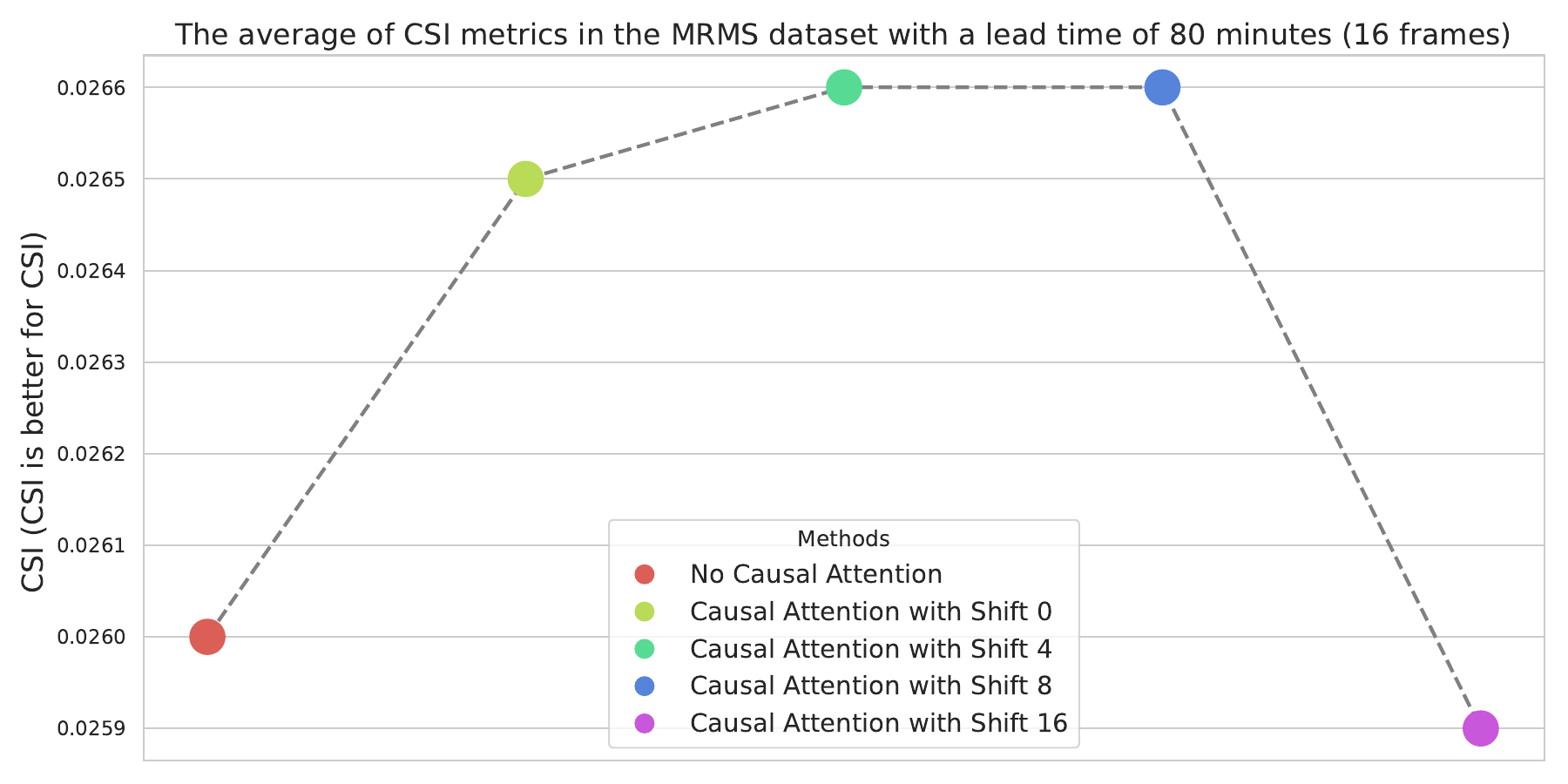}
          \label{fig:csi351}
          }
          \subfigure[CRPS]{
          \includegraphics[width=0.22\textwidth]{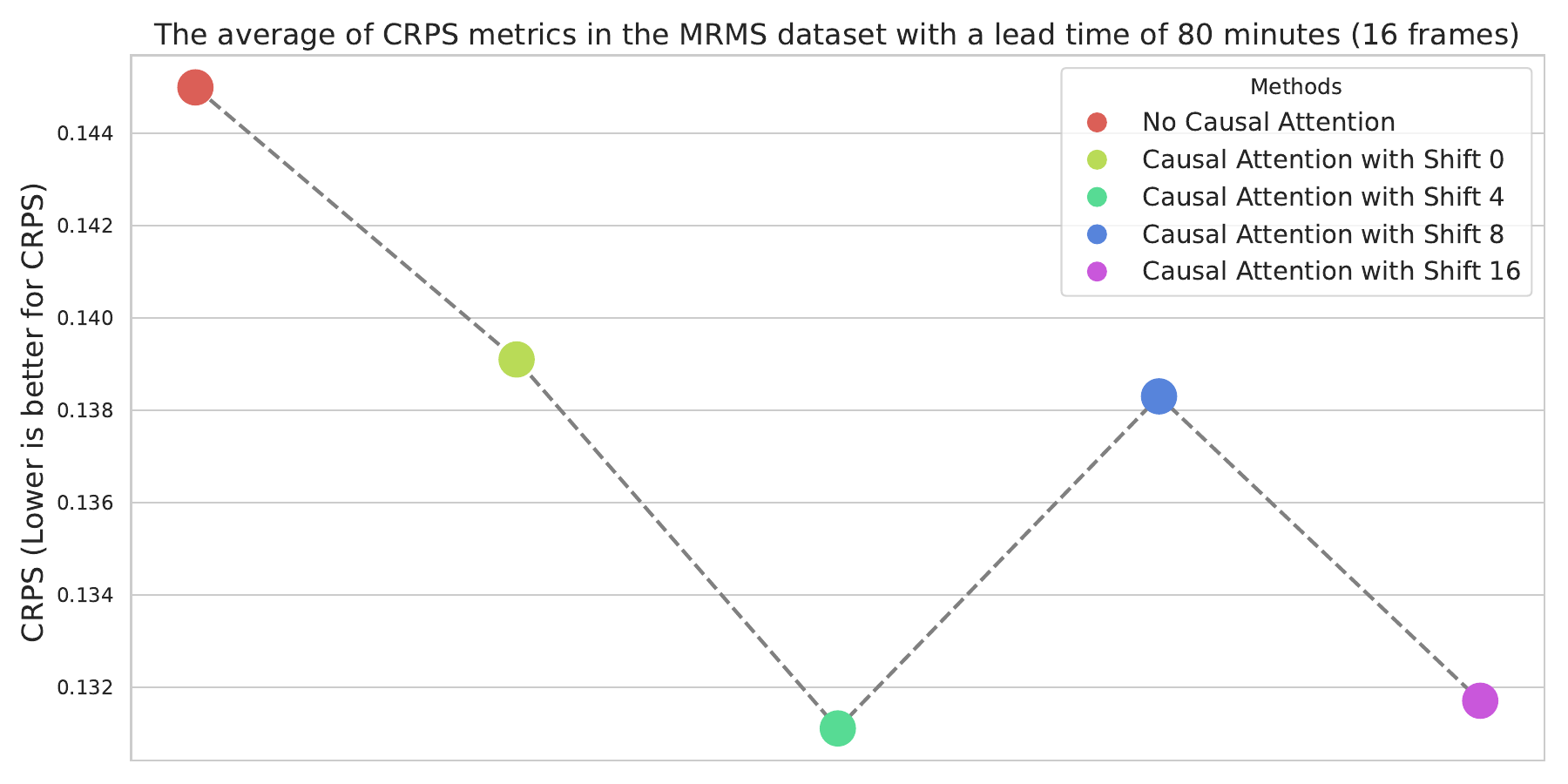}
          \label{fig:crps1}
          }
          \hfill
          \subfigure[FSS]{
          \includegraphics[width=0.22\textwidth]{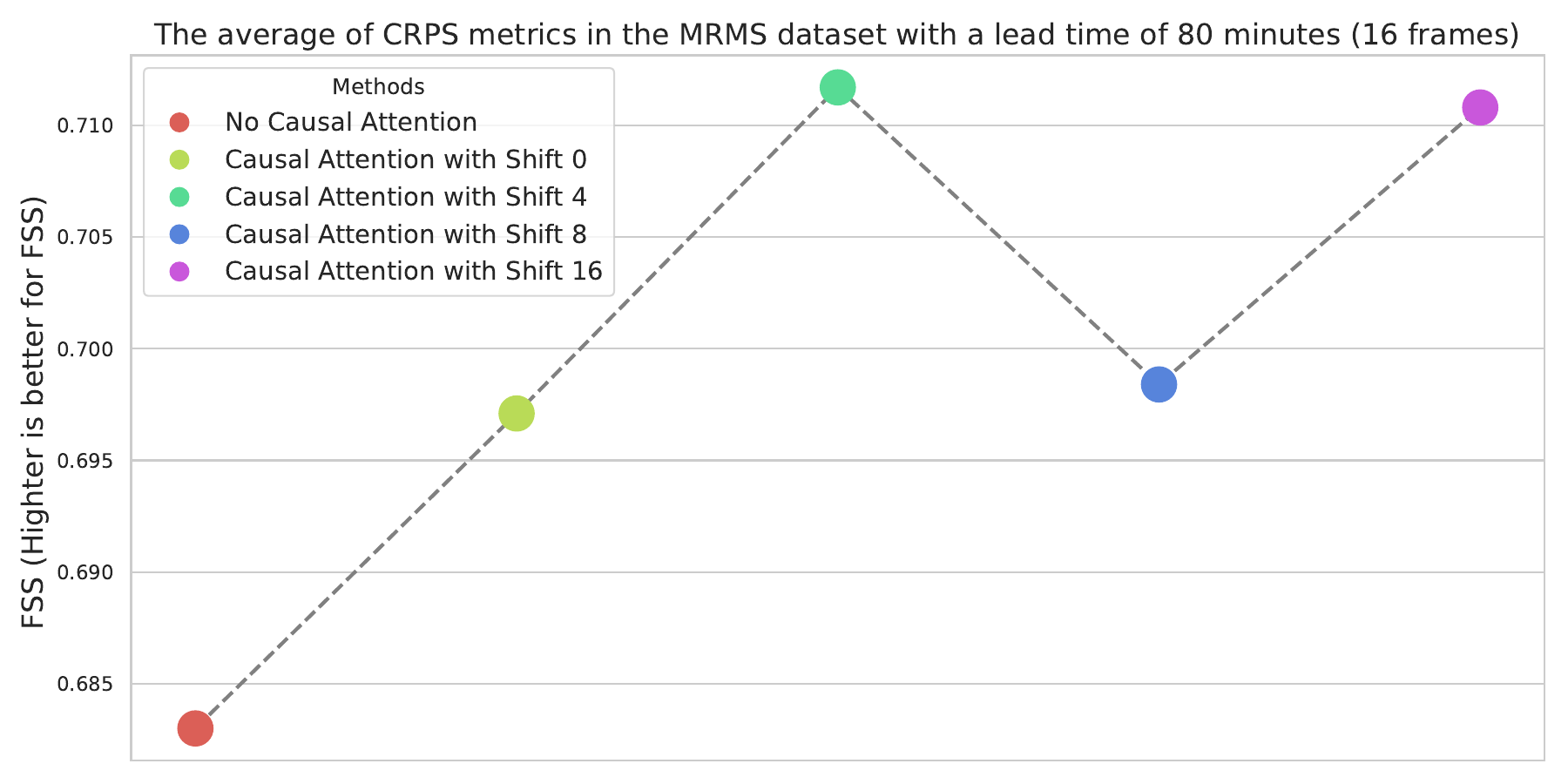}
          \label{fig:fss1}
          }
          \caption{Comparison of precipitation prediction performance for 80 minutes on the MRMS dataset, with and without Causal Attention and different numbers of Channel-To-Batch Shifts.}
          \label{fig:asz1}
          \end{figure*}
  This section will experimentally compare the four variants described in Chapter 2 to determine the optimal spatiotemporal capture scheme. Figure 3 shows the precipitation prediction performance of these models on the SW dataset. The experimental results reveal that, under the same number of training epochs (300), Variant 1 - which calculates attention for all tokens simultaneously - performs best in terms of spatiotemporal feature capture and prediction effectiveness. The data clearly shows that the comprehensive performance ranking is: the entire model calculating spatiotemporal attention simultaneously on all tokens ($F_{JST}$), spatial mixed attention ($H_{JST}+S$), spatiotemporal separated attention ($ST$), and temporal mixed attention ($H_{JST}+T$). This demonstrates the importance of comprehensively considering all information when dealing with complex spatiotemporal data. The superior performance of Variant 1 stems from its ability to more comprehensively capture the interactions and dependencies between temporal and spatial dimensions. In contrast, other variants may miss some key correlations due to separating temporal and spatial information at some stage. Therefore, for the sake of prediction quality, subsequent experiments opted for the JST full spatiotemporal attention approach.

  \subsection{Ablation experiment of Causal Attention mechanism and Channel-To-Batch Shift}

    We investigated the effects of cross-attention mechanism and Channel-To-Batch Shift on short-term precipitation forecasting. We conducted experiments on two datasets, Sweden and MRMS, with a forecast duration of 80 minutes (16 time frames). To comprehensively evaluate model performance, we employed multiple metrics including CRPS, CSI, and FSS.
    From the experimental results on the Sweden dataset (as shown in\figref{fig:csi},\figref{fig:csi35},\figref{fig:crps} and \figref{fig:fss}), we observed that using the cross-attention (Causal Attention) mechanism improved most metrics compared to the baseline model without causal attention. This indicates that the cross-attention mechanism can effectively enhance the model's ability to predict precipitation events. Additionally, we noticed that as the number of batch shifts (0, 4, 8, 16) increased, model performance showed a trend of first improving and then declining. Specifically, when the batch shift was 4, the model achieved optimal performance across most metrics. Compared to the baseline model, the model with a shift of 4 significantly improved the CSI metric (>6.3mm/h) by about 0.02 and the FSS metric by 0.001 at the rainfall threshold.
    Similar trends were observed in the MRMS dataset (results shown in \figref{fig:csi1},\figref{fig:csi351},\figref{fig:crps1} and \figref{fig:fss1}). The model achieved the best overall performance when the batch shift was 4. On the CRPS metric, performance improved by about 0.12; on the FSS metric, it improved by about 0.3; and on the CSI 1 and CSI 8 thresholds, performance improved by 0.2 and 0.01 respectively. These improvements clearly demonstrate the positive impact of the cross-attention mechanism and appropriate batch shifting on enhancing precipitation prediction accuracy.

\section{Conclusion}
In this paper, we propose a Transformer-based diffusion model called DTCA for addressing the complex task of short-term precipitation forecasting. To fully utilize spatiotemporal information, we designed and analyzed four different spatiotemporal modeling variants. Through experiments, we found that allowing all tokens to participate in calculations across spatiotemporal dimensions yields the best predictive performance, highlighting the importance of global spatiotemporal dependencies for accurate forecasting.
Furthermore, we introduced a causal attention mechanism, enabling the prediction process to establish global spatiotemporal dependencies with conditional information. This design effectively leverages conditional information, allowing the model to make more precise predictions about the future based on historical spatiotemporal evolution patterns. Compared to traditional spatiotemporal prediction methods, this mechanism better captures complex spatiotemporal dynamic features, improving the accuracy and reliability of predictions.
To further enhance the model's representational capacity, we proposed an innovative Channel-To-Batch Shift operation. This operation creates "extended samples" by redistributing part of the data from the channel dimension to the batch dimension, thereby enriching feature representation. Significant performance improvements were achieved on two rainfall datasets, demonstrating its effectiveness in enhancing feature representation.
In analyzing the prediction results, we observed the occurrence of some flickering phenomena. This may be attributed to the lack of temporal operations in the decoder. The loss of temporal information during the decoding process potentially affects the continuity and smoothness of predictions. To address this issue, future work could explore high-quality 3D autoencoders. By downsampling in both time and space dimensions, it would be possible to reduce the number of temporal tokens, thereby improving the computational efficiency of the model. Simultaneously, this approach could enhance the capture of hierarchical structure and long-range dependencies in spatiotemporal data.

\bibliographystyle{IEEEtran}

\bibliography{egbib}

\end{document}